\documentclass[11pt, a4paper, twocolumn]{article}
\usepackage{xcolor}
\newcommand{\modif}[1]{\textcolor{black}{#1}}
\usepackage{cite}
\usepackage{amsmath,amssymb,amsfonts}
\usepackage[margin=0.6in]{geometry}
\usepackage{algorithmic}
\usepackage{graphicx}
\graphicspath{{}}
\usepackage{textcomp}
\usepackage{url}
\usepackage{hyperref}
\usepackage{caption}
\usepackage{subcaption}
\usepackage{array}

\begin{document}
\title{Self-mentoring: a new deep learning pipeline to train a self-supervised U-net for few-shot learning of bio-artificial capsule segmentation}
\author{Arnaud Deleruyelle, Cristian Versari, and John Klein} 
\date{Univ. Lille, CNRS, Centrale Lille, UMR 9189 - CRIStAL, F-59000 Lille}

\maketitle

\begin{abstract}
\textit{Background:} 
Accurate segmentation of microscopic structures such as bio-artificial capsules in microscopy imaging is a prerequisite to the computer-aided understanding of important biomechanical phenomenons. State-of-the-art segmentation performances are achieved by deep neural networks and related data-driven approaches. Training these networks from only a few annotated examples is challenging while producing manually annotated images that provide supervision is tedious. 

\noindent\textit{Method:} 
Recently, self-supervision, i.e. designing a neural pipeline providing synthetic or indirect supervision, has proved to significantly increase generalization performances of models trained on few shots. The objective of this paper is to introduce one such neural pipeline in the context of micro-capsule image segmentation. Our method leverages the rather simple content of these images so that a trainee network can be mentored by a referee network which has been previously trained on synthetically generated pairs of corrupted/correct region masks. 

\noindent\textit{Results:} 
Challenging experimental setups are investigated. They involve from only 3 to 10 annotated images along with moderately large amounts of unannotated images. In a bio-artificial capsule dataset, our approach consistently and drastically improves accuracy. We also show that the learnt referee network is transferable to another Glioblastoma cell dataset and that it can be efficiently coupled with data augmentation strategies. 

\noindent\textit{Conclusions:} 
Experimental results show that very significant accuracy increments are obtained by the proposed pipeline, leading to the conclusion that the self-supervision mechanism introduced in this paper has the potential to replace human annotations.
\end{abstract}

\vspace{1em}
\noindent\textbf{Keywords}: 
deep learning, few shot learning, microscopy image segmentation, self-supervised learning, semi-supervised learning, U-net.

\maketitle

\section{Introduction}
\label{sec:introduction}


Studying the mechanical properties of living cells is an essential step in the understanding of macroscopic effects on the functioning of some organs or species. In particular, it is important to understand how cells flow inside a viscous fluid contained in a network of vessel-like structures. 
It is, however, an extremely challenging task because like any biological matter, cells are subject to many intrinsic or environmental factors of variation. The levels of elasticity and shape are examples of such intrinsic factors and viscosity level and pipeline shape are examples of environmental ones. 
For the case of blood cells, and in order to keep those factors at bay and acquire an in-depth understanding of the mechanical properties of this class of cells, it is possible to create micro-capsules with similar intrinsic properties as standard blood cells and observe how they flow into a controlled (and reproducible) environment whose main constituent is a micro-channel containing a fluid with known specifications. 

Thanks to recent advances in micro-fluidics \cite{sevenie2015characterization}, it is possible to acquire good quality images of micro-capsules flowing inside a channel with a $100 \times 100 \mu \textrm{m}^2$ square section. To derive the sought mechanical properties (deformation, velocity), it is necessary to obtain, for each acquired image, a micro-capsule segment, i.e. a binary image where the positive pixels occupy an image region corresponding to the position of the analyzed capsule. 
Automating this task is an image processing problem known as segmentation and the solution of this problem is a function that maps input images to masks.

In the past decade, supervised learning methods relying on deep neural networks have considerably improved segmentation performances compared to prior arts that were either relying on non-data-driven models or on older generation of machine learning models. In particular, an architecture called U-net \cite{ronneberger2015u} has become the default model to choose. However, training U-net requires a rather large amount of annotated data to achieve the highest level of segmentation performances. 

Unfortunately, image segmentation is probably one of the tasks for which producing annotated images is the most tedious. Indeed, this supervision is usually obtained from a domain expert that will use a delineation software to manually identify the pixels that belong to the object contour. Not only do those experts have limited time for annotation but also this activity is patently tiresome. 

Based on this observation, this paper investigates the possibility to use side information in order to train a U-net architecture efficiently from a handful of labeled data. 
Our method consists in training several U-nets that will play different roles as part of a global learning scenario. A first U-net, called referee, is trained to map an incorrect capsule mask to a correct version of it. 
Then, a second U-net, called trainee, is pretrained on the small annotated dataset. 
In a subsequent training episode, the trainee and the referee can be plugged together so that the trainee is asked (through a corresponding loss term) to produce mask regions that will not be modified by the referee, meaning that they deserve no correction. 
An important advantage of this approach is that the referee can be trained on easy to produce synthetic input/output pairs. Indeed, from a single capsule mask of the labeled dataset, one can apply many arbitrary deformations that the referee will learn to cancel. 
Also, the conjunction of these two networks with a third one allows to exploit unlabeled object images, making it compliant with the so-called semi-supervised paradigm. This is particularly interesting as unlabeled images are not difficult to obtain. 
To exploit those images, the third U-net is trained to reconstruct the original image from a binary mask. This third network helps the referee and trainee to avoid learning undesirable trivial mappings such as a mapping from object images to all-black masks which the referee would regard as a prediction that does not deserve corrections. 

The next section presents a number of related works on segmentation and strategies to train neural networks from a limited amount of labeled data. The third section provides necessary definitions and formalizes concepts that are the building blocks of our approach which is detailed in section \ref{sec:contrib}. Our training pipeline is validated on a recent micro-capsule dataset \cite{Caps2021} in section \ref{sec:exp} and improves segmentation accuracy by a large margin. Another cell dataset is also used to show that the contribution can be applied to a broader spectrum of microscopy image segmentation. Finally, a discussion and conclusion summarizes our contributions and gives some perspectives for additional developments.


\section{Related works} 
\label{sec:related_works}


This paper focuses on semantic segmentation in which all objects (micro-capsules in our bio-mechanical context) appearing in one image are meant to be mapped to a single common binary mask. Instance segmentation in which each object instance is mapped to a different mask is out of the scope of this article. 
Non-data-driven approaches assume that the region to detect exhibits a form of homogeneity, i.e. it is either a uniform region of the image or it can be mapped to one such using appropriate embedding. 
Texture features \cite{chowdhary2020segmentation} or mean-shift filters \cite{comaniciu2002mean} are typically good candidates to achieve such an embedding. Based on this homogeneity assumption, the segmentation problem can be solved using a clustering \cite{beucher1992watershed,chuang2006fuzzy,feudjio2013automatic,chen2018survey}, split-and-merge \cite{horowitz1976picture} or region growing \cite{yu2008irgs} algorithm. Another possibility is to reshape the segmentation problem into an optimization one in which a functional has to be minimized \cite{mumford1989optimal,chan2001active,chan2006algorithms,pascal2018joint,foare2019semi}. This functional usually comprises a term that approximates the image with a piecewise smooth function and terms that penalize rapid variations of the approximation inside smooth regions as well as the contour length of regions. When it comes to object segmentation, the limitation of these data-free approaches is obviously the homogeneity assumption because an object is most often a non-homogeneous region. Applying those methods in our context would typically lead to over-segmented regions, sometimes also called superpixels \cite{ren2003learning}, which would have to be aggregated using a second additional inter-region model.

In the past decade, deep neural networks have proved to be able to perform segmentation directly at the object level. The most popular architecture in this vein is called U-net \cite{ronneberger2015u}. It is a fully convolutional network (FCN \cite{long2015fully}) with an encoder-decoder structure. The input image is down-sampled by the encoder and up-sampled by the decoder. Skip connections between layers that process feature maps of identical sizes allow to mitigate vanishing gradient issues during backpropagation. A similar architecture known as SegNet \cite{badrinarayanan2017segnet} shares many ideas with U-net but uses an up-sampling layer that does not have trainable parameters. 
Building upon such FCN architectures, some approaches on sequences of natural images achieve impressive results \cite{caelles2017one} but they are pre-trained on massive supervised datasets which does not match our effort to escape high levels of supervision. Another interesting achievement is proposed in \cite{el2020bb,tian2021boxinst} where semantic segmentation is achieved from imperfect supervision (bounding box of the object instead of mask) but again this is a different working assumption than the one tackled in this paper. Some authors also proposed to plug recurrent layers \cite{lin2018multi} or attention modules \cite{li2020attention} in FCNs which improves segmentation performances. 

The achievements obtained by deep neural networks on segmentation tasks cannot be denied, especially for cell microscopy images \cite{kiran2022denseres,hamida2021deep,zhao2022lfanet,lal2021nucleisegnet} which is the focus of this article.  
In spite of these achievements, deep neural networks suffer from a strong need of massive amount of supervised data. 
Consequently, the machine learning community started to investigate workarounds that would allow to learn from limited amount of data. 
As proposed in \cite{yu2022multiheadgan} for the specific context of retinal pigment epithelium cell segmentation, a clever way to learn from limited annotated data is for example to enforce consistency in the trained model through a more general-purpose generative objective function. 
This approach was developed in a context where around 150 annotated images and ten times more unlabeled data were available. 
The situation where the number of labeled data is far smaller (typically around a dozen) is known as few shot learning (FSL) and is the framework where we place ourselves in this paper. 
Data augmentation (DA) \cite{shorten2019survey} consists in applying various transforms (e.g. rotation, symmetry, noise injection) to a strain input in order to create additional inputs. If supervision is available for strain inputs, we obtain new input/output pairs that can somewhat close the gap between usual supervised learning and FSL. 
DA applied to labeled data only is not a general answer to the FSL setting because images in the test phase will still be significantly different from those seen during training. 
In this paper, while the investigated solution remains compatible with DA (as will be shown in the experimental section) for labeled data, we focus on solutions working for a fixed amount of such data. 

Another working assumption is that we assume having access only to very few labeled data and a moderately large amount of unlabeled data but all these data are solely dedicated to the addressed segmentation task which rules out multi-task and meta-learning approaches \cite{finn2017meta,snell2017prototypical,ren2018meta,tian2020rethinking,dawoud2020few} that are also instrumental for FSL.

A powerful solution to address our semi-supervised FSL setting is self-supervised learning (SSL). SSL has gained considerable attention in the past few years and consists typically in leveraging the structure of unlabeled data to obtain supervisory signals for free in order to guide the model into learning meaningful representations. 
To obtain such a signal, one can minimize additional input reconstruction loss terms. 
This idea was already exploited in workflows that learn an embedding function from unsupervised data in order to find a lower dimensional representation and then plug this pre-trained embedding function in a model solving the supervised task \cite{erhan2010does}. 
Auto-encoders \cite{ranzato2006efficient,bengio2006greedy} and siamese networks \cite{bromley1993signature,chen2021exploring} are relevant candidates to learn meaningful embeddings and thus play an important role in SSL. 
Coupled with certain forms of DA, denoising auto-encoders \cite{vincent2010stacked} are meant to reconstruct the input from corrupted versions of it. 
The same idea is also one part of the success of attention models such as BERT \cite{devlin2018bert} which uses masked inputs. In natural language processing tasks, asking the model to predict a missing word inside a sentence provides powerful feedback for learning.

Instead of learning to fill gaps, it is also possible to replace a word with another (unrelated) one and ask the model 
to learn that the correct and incorrect sentences should be far away after embedding. More generally, this idea is known as contrastive SSL \cite{hadsell2006dimensionality} which also uses positive input pairs (inputs that are similar and should be close to each other after embedding). 

In this article, we propose to use another form of SSL known as self-training \cite{xie2020self} that has strong connection with knowledge distillation \cite{44873} and involves two different networks: a student network and a teacher network. While the initial motivation behind knowledge distillation is to obtain a compact model from an ensemble or a very large model, self-training leverages knowledge distillation as a way to provide supervisory signals. 
In self-training, a teacher network is first trained on the annotated images. The teacher is then used to produce pseudo-annotations for the non-annotated images. The student network is then trained using both the true and pseudo labels. As proposed in \cite{xie2020self}, this process can be coupled with DA and it is possible to iterate by using the trained student as a teacher for another to-be-trained student. In \cite{pham2021meta}, the authors propose to alternate between student and teacher optimizations so that the teacher can adjust its pseudo-labels in order to maximize the student performance.

An innovative aspect of our contribution is that we depart from this pseudo-label generation idea and, instead, we use a trainee network
that is meant to solve the main segmentation task and a referee network that maps segmentation masks issued by the trainee to corrected versions of them.  
Indeed, denoising SSL can be used to train the referee. 
Because the structure of binary masks is extremely simple, we can generate a very large amount of synthetic masks to train a strong referee. Once the referee model is trained, consistency regularization will force the trainee to minimize the distance between its outputs and their corrected (referee approved) versions. 
In \cite{laine2016temporal}, this concept is used for a stochastic (dropout) network. The trainee and referee are different realizations of the network for each batch and are required to produce similar predictions for inputs obtained from randomized DA. The referee can also be obtained as a running prediction average across epochs. 
In \cite{tarvainen2017mean}, a running average of the trainable parameters of the referee is also performed. 
Unlike these approaches, in our case, the referee and the trainee solve different tasks thereby transferring to the trainee a different kind of knowledge. Consequently, we refer to our approach as self-mentoring. 

From the above analysis of the literature, the originality of our contributions can be summarized as such:
\begin{itemize}
    \item we propose a segmentation methodology for training any deep neural networks from a very few labeled data, 
    \item we do not require external data or any form of transfer learning obtained from other datasets or tasks,
    \item we exploit the structure of binary mask images to train a self-supervised network that will transfer knowledge to the one used for segmentation. 
\end{itemize}

It should also be clarified that we do not address fully self-supervised segmentation \cite{wang2022fully} where no label is available and segmentation performances are consequently far more limited.

\section{Background on deep learning based image segmentation}
\label{sec:basics}

\subsection{Problem statement} 
\label{sub:generalities_and_background}

In the semi-supervised setting addressed in this paper, one has access to a training dataset $\mathcal{S}_{\text{tr}}$ which contains $n_{\text{s-tr}}$ pairs $\left( \mathbf{x}^{(i)}, \mathbf{y}^{(i)} \right)  $ of inputs/targets, for $i \in \left\{ 1.. n_{\text{s-tr}} \right\}$. Another smaller set $\mathcal{S}_{\text{val}}$ which contains $n_{\text{s-val}}$ pairs will be used to detect model convergence. 
In our segmentation context, the vector $\mathbf{x}^{(i)}$ is an image of a micro-capsule inside some channel and vector $\mathbf{y}^{(i)}$ is a binary image with same size as $\mathbf{x}^{(i)}$ and corresponds to the binary mask of the region occupied by the capsule that can be seen in $\mathbf{x}^{(i)}$. 
Images $\mathbf{x}$ are supposed to be standardized so that pixel values are in $\left[ 0;1 \right] $. For the binary images $\mathbf{y}$, pixel values are in $\left\{ 0;1 \right\}$. We also specifically place ourselves in a setting where images $\mathbf{x}$ have a simple structure in the sense that they can be described by only two regions: the one occupied by the object to segment and a uniform background. The visual features of each region type should be similar for two different input images with at most only one object to segment per image or no object at all. These assumptions are compliant with the micro-capsule dataset that is the focus of this paper but also for a number of microscopy image datasets which are acquired with a high level of control and exhibit very limited visual disparity.

Moreover, one has also access to an unsupervised training set $\mathcal{U}_{\textrm{tr}}$ and an unsupervised validation set $\mathcal{U}_{\textrm{val}}$ which contains respectively $n_{\text{u-tr}}$ and $n_{\textrm{u-val}}$ input images for which no annotation is available. Furthermore, we assume $\mathcal{S}_{\text{tr}}$ is small and typically contains no more than ten images (and even fewer in $\mathcal{S}_{\text{val}}$). The number of unlabeled images is larger by several orders of magnitude $n_{\text{u-tr}} + n_{\text{u-val}}\gg n_{\text{s-tr}} + n_{\text{s-val}}$. Since labeled data can always be stripped out of labels to provide unlabeled data, the input images contained in $\mathcal{S}_{\text{tr}}$ are also in $\mathcal{U}_{\text{tr}}$. Similarly, input images in $\mathcal{S}_{\text{val}}$ can be used in $\mathcal{U}_{\text{val}}$. However, input images in $\mathcal{S}_{\text{tr}}$ cannot be used in $\mathcal{U}_{\text{val}}$.

In the approach introduced in this paper, several networks will be used. Each of them is a parametric model. The one that solves the main segmentation (at the end of the whole training process) is the trainee neural network $f_{\boldsymbol\theta}^{(\text{tne})}$ in the sense that it maps inputs to predicted region masks. The trainee will be mentored by the referee network $f_{\boldsymbol \phi}^{(\text{ref})}$ which maps imperfect region masks to higher quality region masks.
Finally, the reverse-pipeline network $f_{\boldsymbol \varphi}^{(\text{rev})}$ maps region masks back to inputs. All of them are three instances of the U-net architecture which is presented in more details in the next paragraphs.


\subsection{U-net architecture} 
\label{ssub:U-net_architecture}

As mentioned before, the dominating approach in the literature for training a model for segmentation are deep neural networks with a U-net architecture \cite{ronneberger2015u}. Even if the training process introduced in this paper can be applied to any architecture designed for segmentation, we will use a U-net architecture for each of the three networks. 

The U-net architecture contains two modules called the encoder and the decoder. Like many other encoding/decoding architectures \cite{ranzato2006efficient,vincent2010stacked}, the motivation behind this is to learn a lower dimensional embedding of the input image that creates a bottleneck which forces the network to filter out irrelevant information in order to achieve the prediction goal. The encoder is a classical fully convolutional module, i.e. a series of $M$ sub-modules each containing a few convolutional layers followed by one max-pooling layer. The decoder is organized in a symmetric fashion and also possesses $M$ sub-modules. However, in the sub-modules of the decoder, the max-pooling layer is replaced with an upsampling layer. In \cite{ronneberger2015u}, the authors use transpose convolution for upsampling which is parametric and pads the input tensor with sufficiently many zeros so that the output one has the desired dimensions. Because our networks are meant to be used on high resolution images, we will use instead a parameter-free upsampling layer that simply performs nearest neighbor interpolation.

\begin{figure*}[!t]
\includegraphics[width=\textwidth]{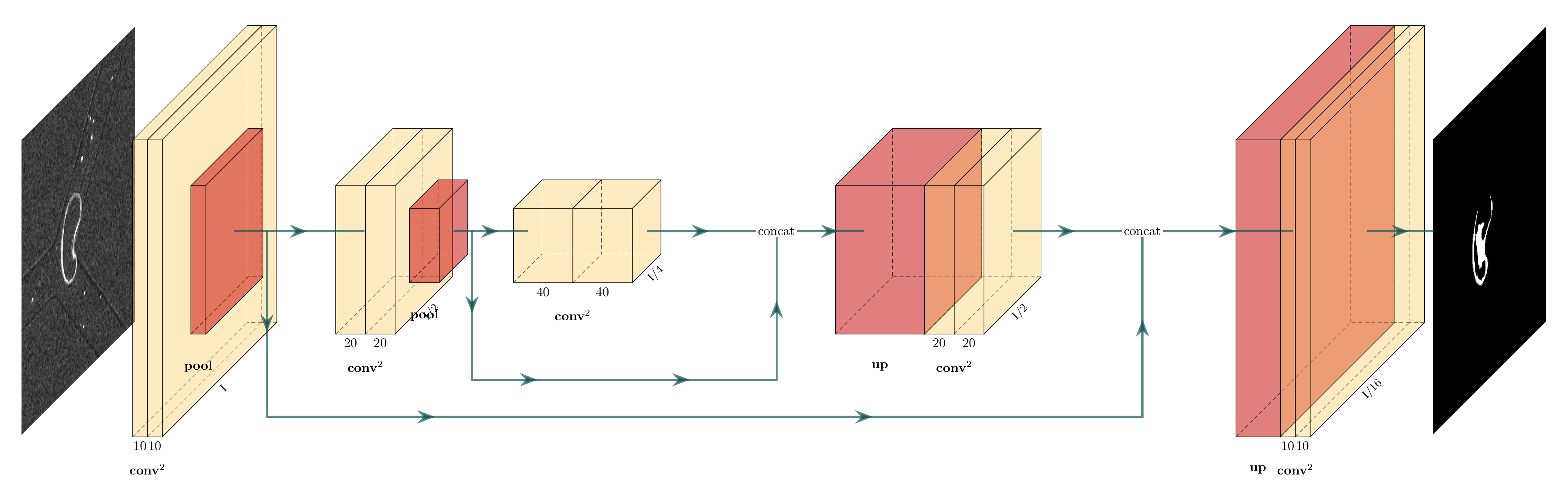}
\caption{U-net architecture: output tensors of 2D convolutional layers are in yellow while output tensors of pooling or upsampling layers are in red. There are $M=3$ sub-modules in both the encoder and decoder. Each sub-modules contains two convolutional layers denoted as $\textrm{conv}^2$. \label{fig:unet}}
\end{figure*}

Figure \ref{fig:unet} provides an illustration of the architecture of U-net for $M=3$ and two series of convolution inside each sub-module. These architectural aspects are shared by all implemented networks in the sequel. They also share filter sizes ($3\times 3$) and pooling sizes ($4\times 4$). 
The number of filters per convolutional layer usually increases as the height and width of tensors decrease. In this paper, we start with $F$ filters in the first layer of the encoder. The number of filters is then multiplied by a factor 2 in each sub-module in the encoder. The decoder works in opposite way and decreases the number of filters by a factor 2 in each sub-module. In Fig. \ref{fig:unet}, this hyperparameter is set to $F=10$. 
Note that convolution layers use padding in order to preserve height and width of tensors. 

\section{Methods}
\label{sec:contrib}

In this section, we introduce our self-supervised learning pipeline for semi-supervised datasets with few labeled samples which we call self-mentoring. We start with a presentation of the micro-capsule segmentation data for which self-mentoring is designed. It is followed by an overview of the general organization in different steps of the self-mentoring pipeline and then, each step is presented in more details. 
 

\subsection{Data} 
\label{sub:data}


\begin{figure*}
     \centering
         \includegraphics[width=.2\textwidth]{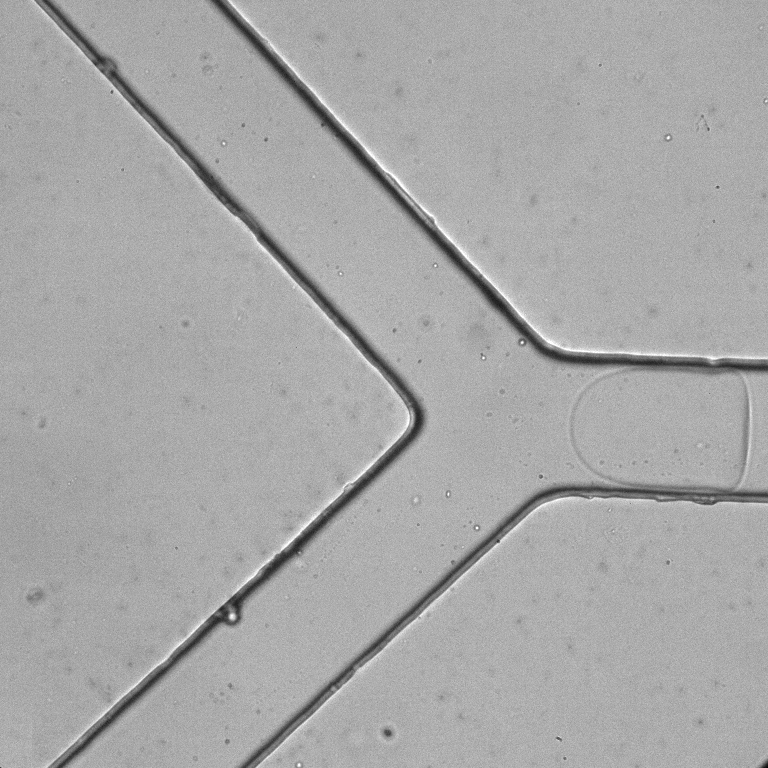}\:
         \includegraphics[width=.2\textwidth]{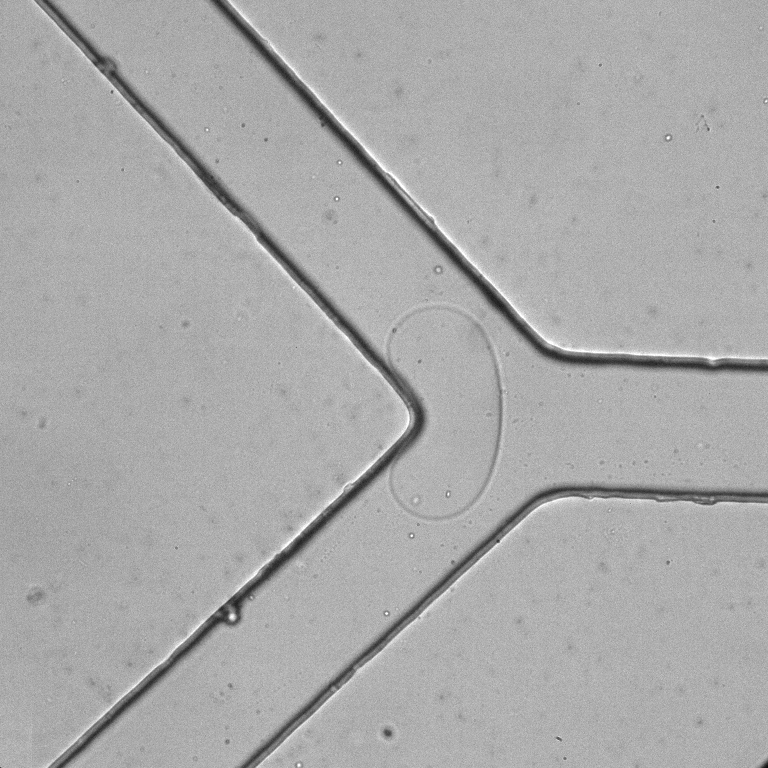}\:
         \includegraphics[width=.2\textwidth]{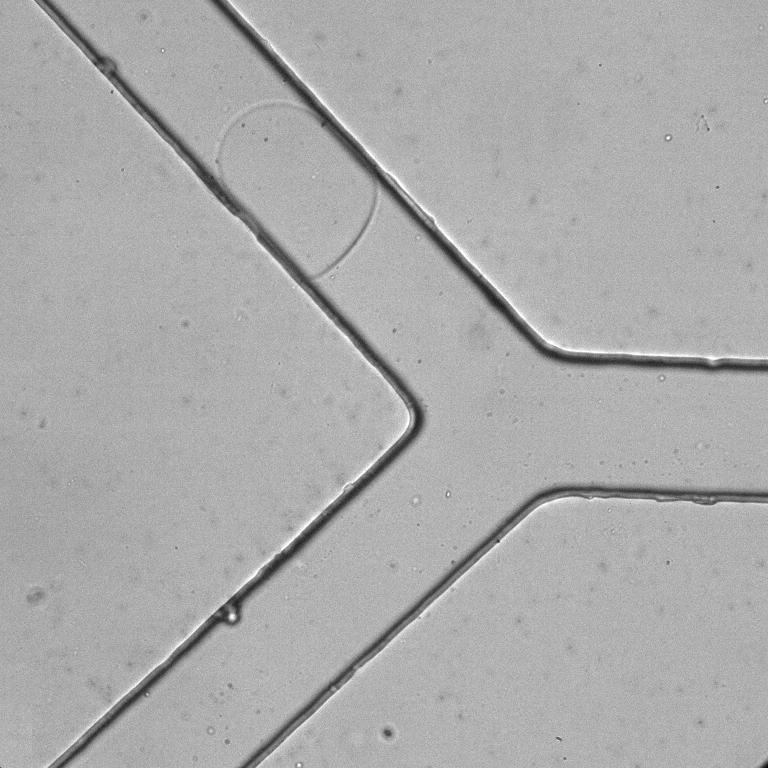}\:
         \includegraphics[width=.2\textwidth]{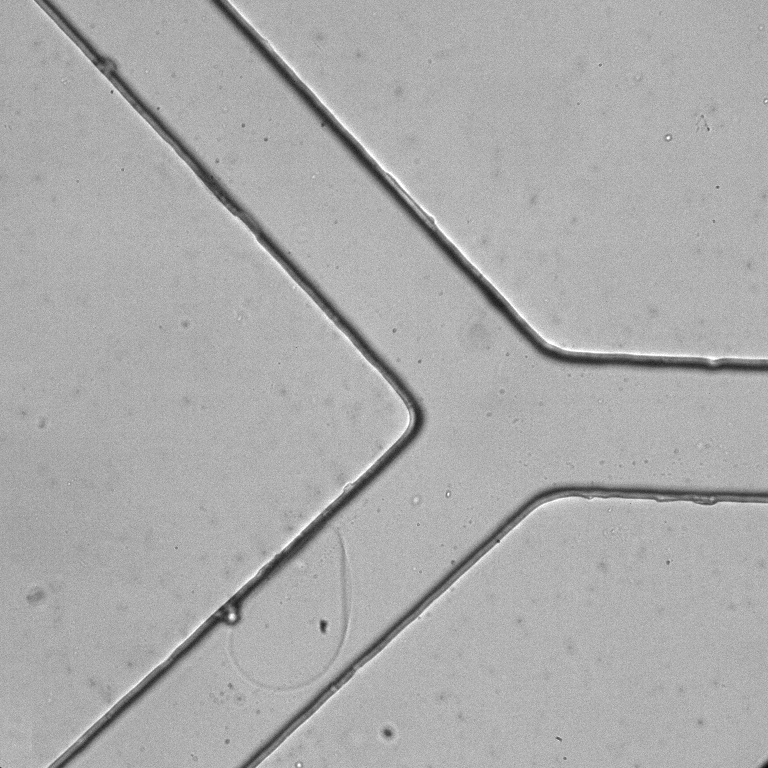}\\
         \includegraphics[width=.2\textwidth]{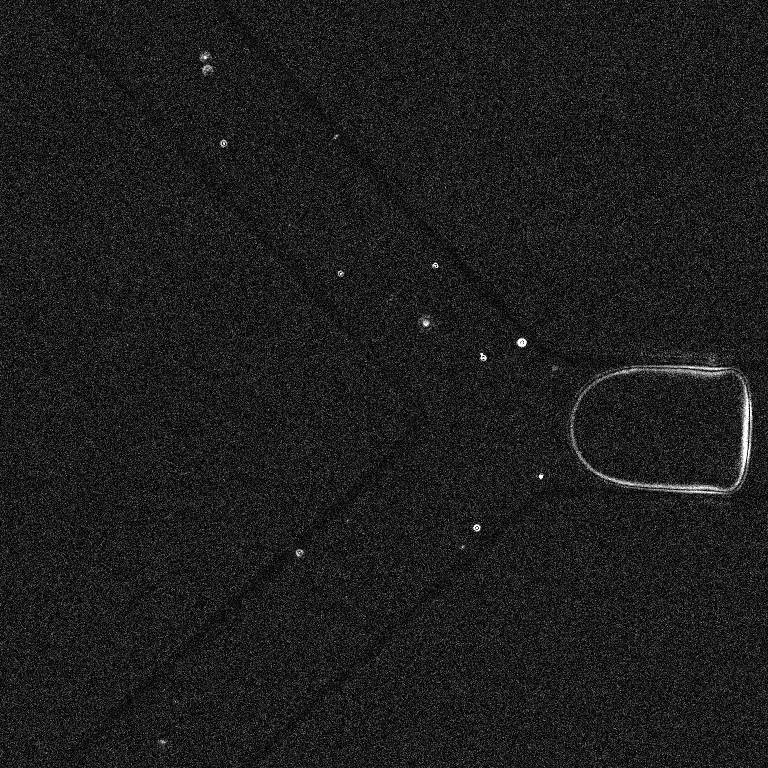}\:
         \includegraphics[width=.2\textwidth]{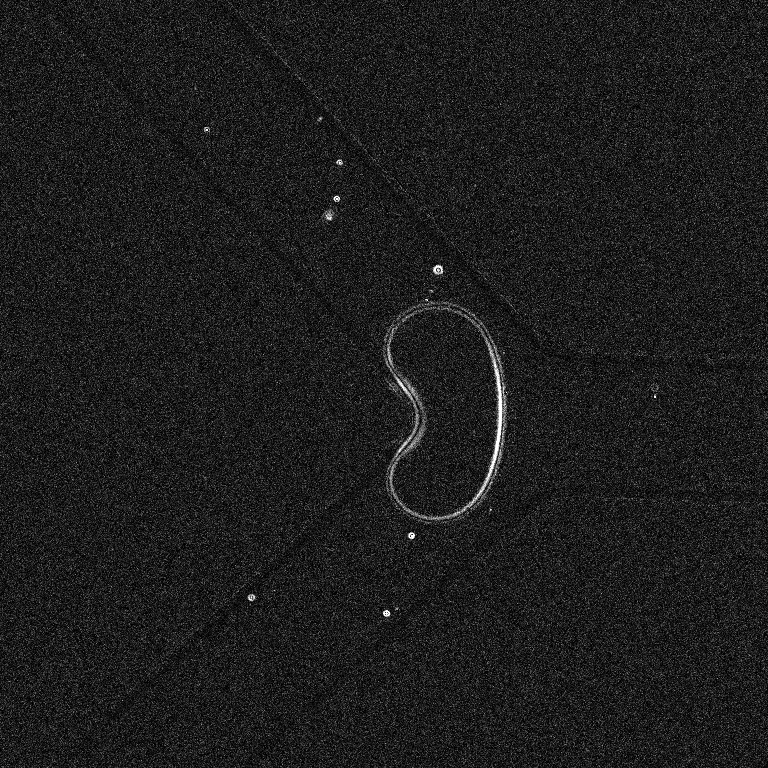}\:
         \includegraphics[width=.2\textwidth]{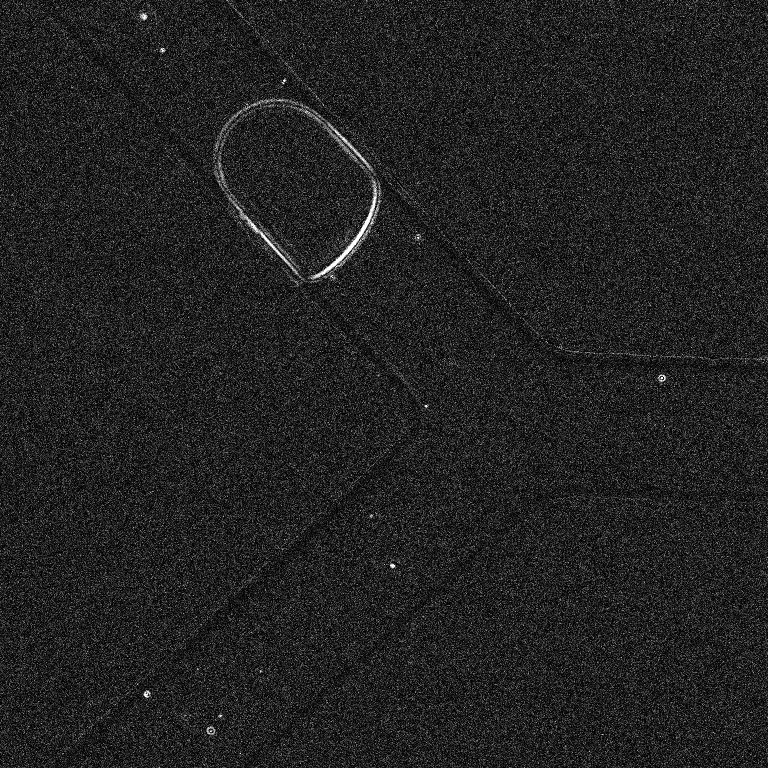}\:
         \includegraphics[width=.2\textwidth]{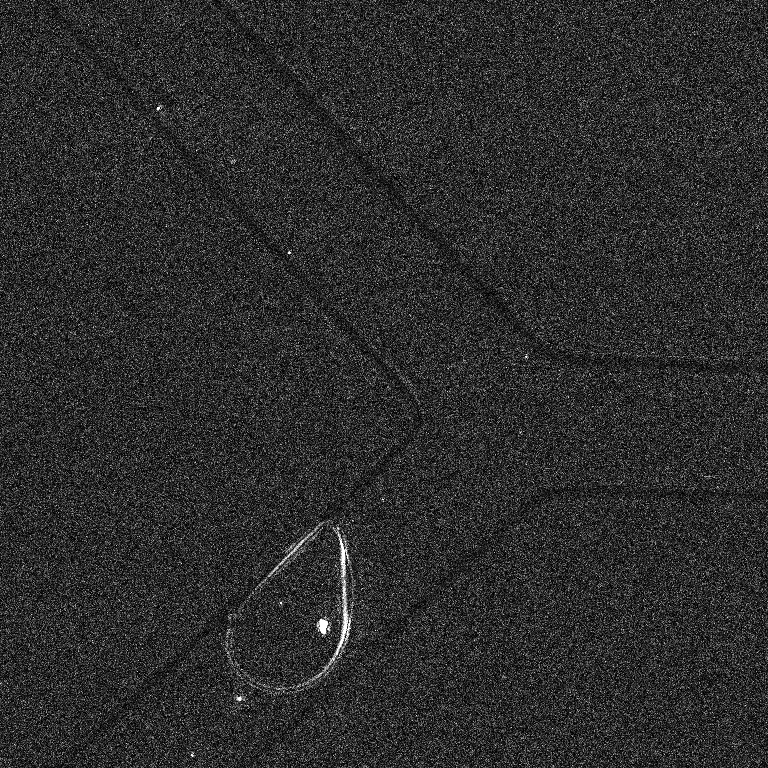}\\
         \includegraphics[width=.2\textwidth]{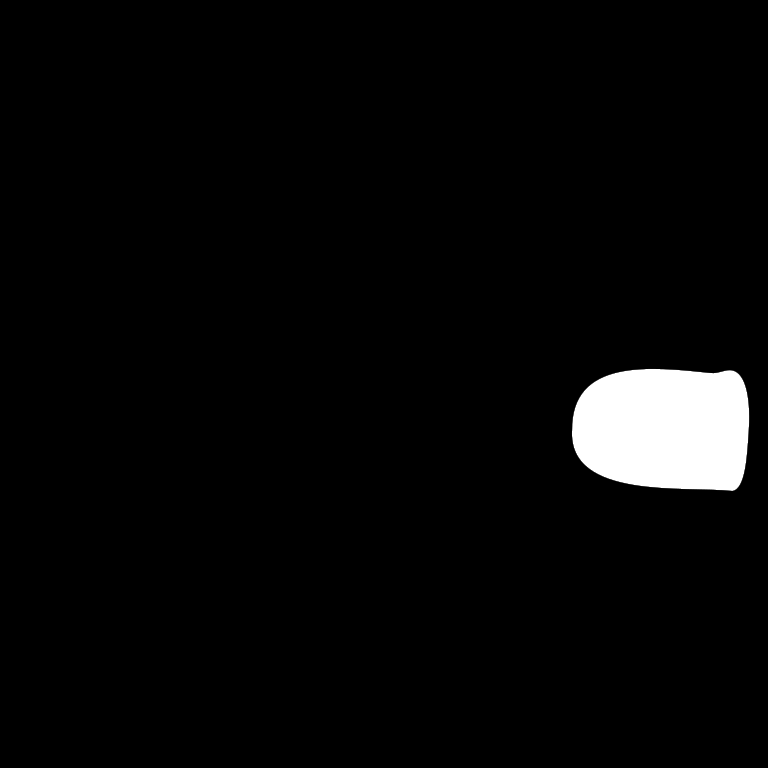}\:
         \includegraphics[width=.2\textwidth]{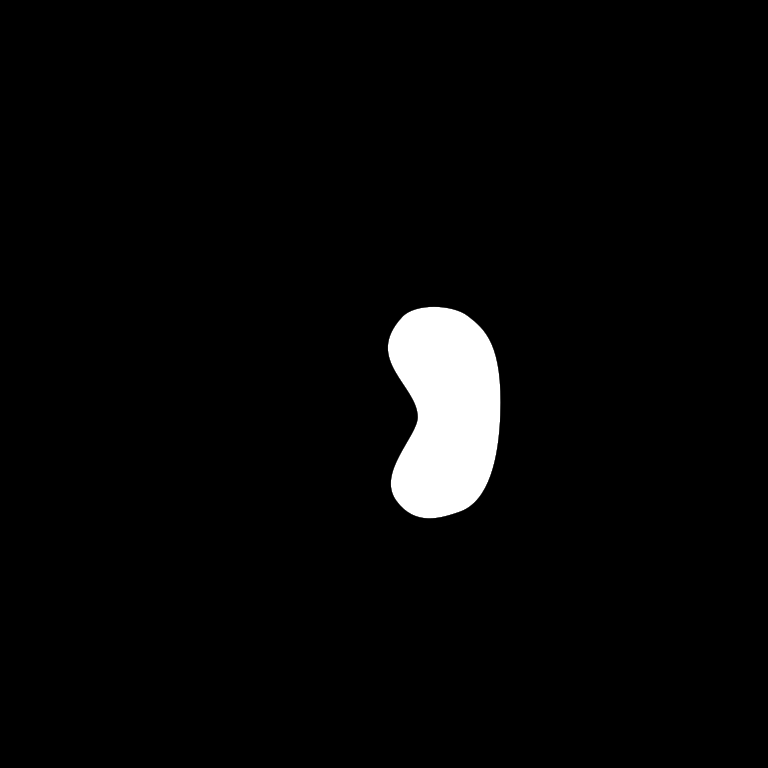}\:
         \includegraphics[width=.2\textwidth]{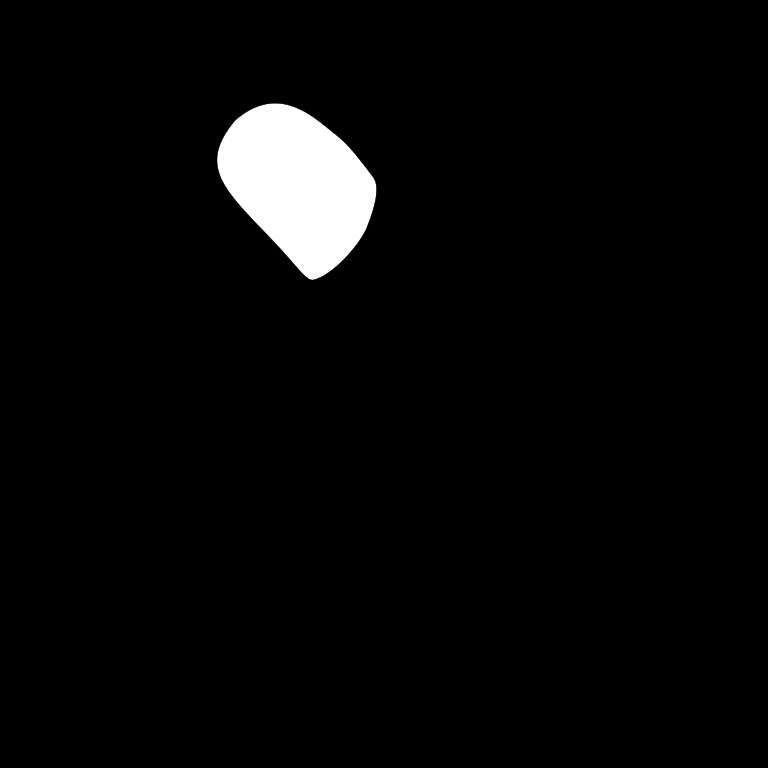}\:
         \includegraphics[width=.2\textwidth]{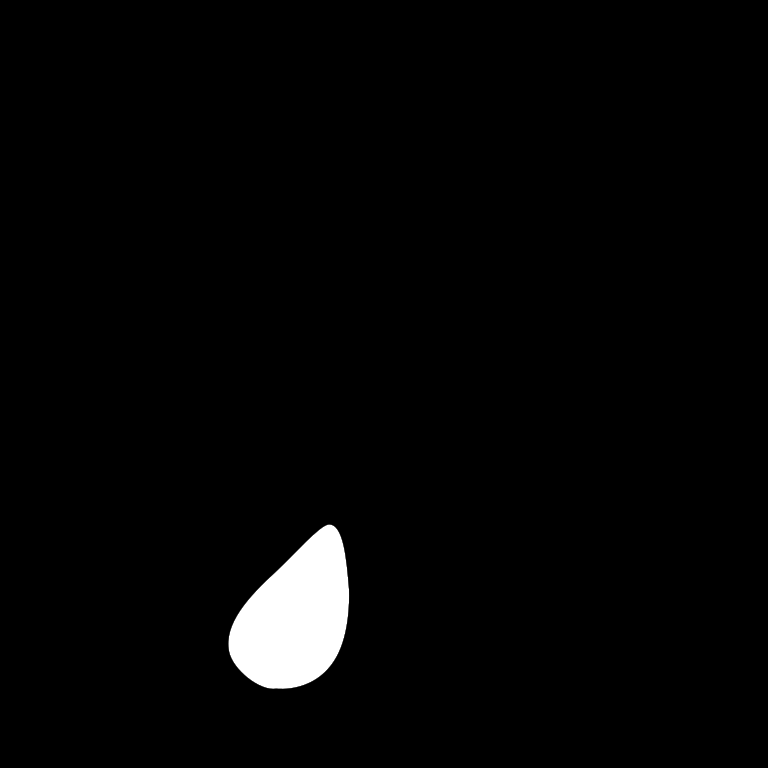}

        \caption{Example of micro-capsule images: raw images (first row), images after background suppression used as inputs $\mathbf{x}$ (middle row), capsule masks $\mathbf{y}$ (final row).}
        \label{fig:caps} 

\end{figure*}

Biomechanics is a field of biomedical engineering that focuses on understanding the mechanical properties of biological structures such as cells. In particular, how cells flow through blood vessels and how this flow impact on cells (deformation, collisions) is an open problem. To better understand these micro-mechanics phenomenons, a possibility is to use artificial capsules (with similar mechanical properties as cells) flowing through a canal and to acquire images of the flow in a controlled acquisition environment. 

The Capsule \cite{Caps2021} dataset is a segmentation dataset which was obtained in this way. It contains 353 labeled grayscale images of $768 \times 768$ pixels. Each image $\mathbf{x}$ shows one capsule flowing through a Y-shaped micro-canal. The micro-canal has a $100 \times 100 \mu \textrm{m}^2$ square section and micro-capsules, which are initially spherical, have a diameter of the same order of magnitude. Images were obtained using a high-speed camera mounted on a microscope. Under the combined effect of confinement and viscous stress, the capsules are deformed. In most images, they exhibit a "bullet" shape but when they are in the vicinity of the channel bifurcation, the deformation is stronger and they exhibit a more elongated shape. Micro-capsules are transparent objects therefore, only their contour are visible in image, as can be seen in Fig. \ref{fig:caps}, first row. Depending on the level of deformation, the contour is quite dim and the pattern to be detected by neural networks is a rather weak signal. To amplify this pattern, a background suppression pre-processing is applied to all images. It simply consists into computing the (pixelwise) median image and subtract that latter to all raw images, see Fig. \ref{fig:caps}, second row for examples of such pre-processed images. Observe that this pre-processing step is unsupervised as it uses input images only.  

An accurate segmentation mask $\mathbf{y}$ is available for each input image, see Fig. \ref{fig:caps}, final row. However, the intent behind the present paper is to avoid having to manually create so many annotated images to analyze future micro-capsule acquisitions. Consequently, our goal is to learn a predictive model that maps $\mathbf{x}$ to $\mathbf{y}$ from only a handful of annotated images so that future acquisitions can be analyzed much faster. A large majority of these masks will thus not be revealed to the models to be trained.

\subsection{Overview} 
\label{sub:overview}

The self-mentoring learning scenario introduced in this paper can be decomposed in four learning phases as follows:
\begin{enumerate}
    \item Train the referee $f_{\boldsymbol \phi}^{(\text{ref})}$ once for all using synthetic region mask images,
    \item Pre-train the trainee $f_{\boldsymbol\theta}^{(\text{tne})}$ using $\mathcal{S}_{\text{tr}}$ and $\mathcal{S}_{\text{val}}$,
    \item Train the reverse net $f_{\boldsymbol \varphi}^{(\text{rev})}$ as well using $\mathcal{S}_{\text{tr}}$ and $\mathcal{S}_{\text{val}}$,
    \item Further train $f_{\boldsymbol\theta}^{(\text{tne})}$ only using $\mathcal{U}_{\text{tr}}$, $\mathcal{U}_{\text{val}}$, $\mathcal{S}_{\text{tr}}$, $\mathcal{S}_{\text{val}}$ and feedback provided by $f_{\boldsymbol \phi}^{(\text{ref})}$ and $f_{\boldsymbol \varphi}^{(\text{rev})}$.
\end{enumerate}

\begin{figure*}[h]
\includegraphics[width=\textwidth]{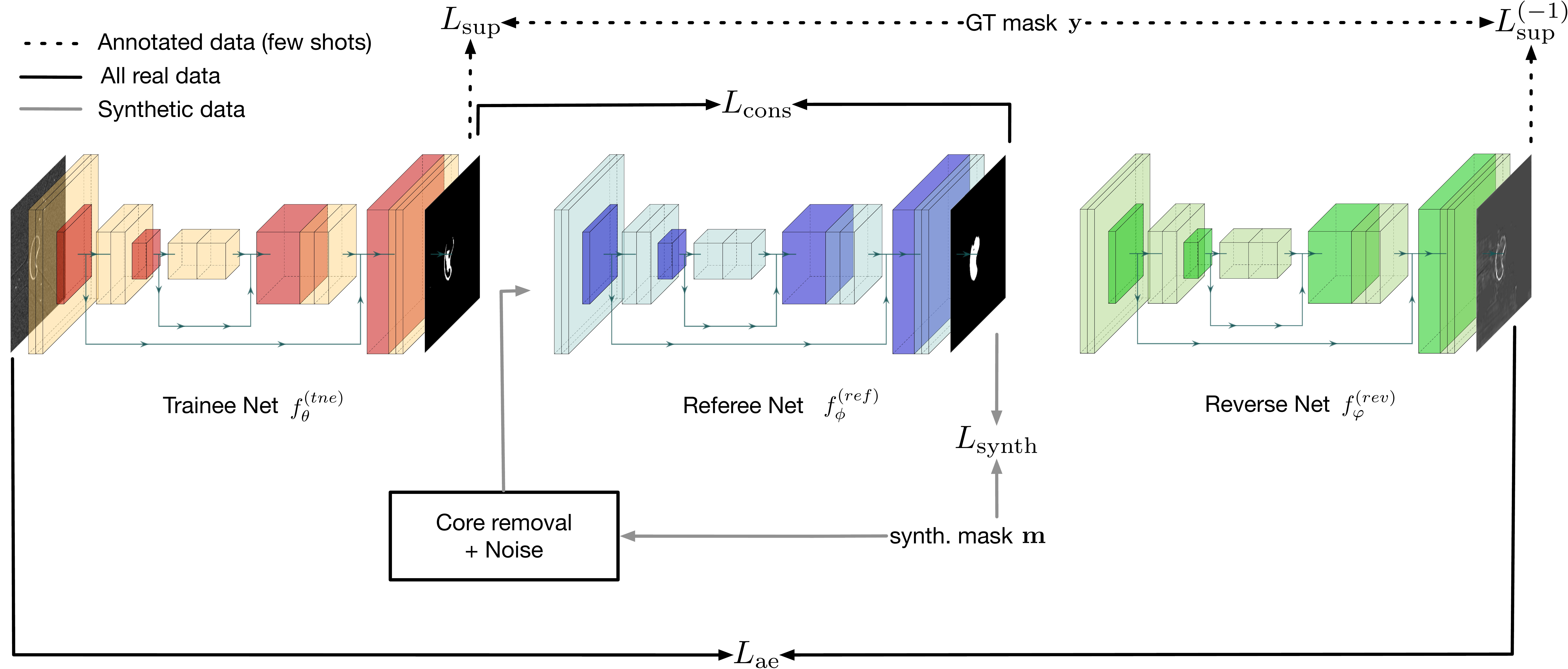}
\caption{Overview of the learning pipeline. An input $\mathbf{x}$ is first processed by the trainee, whose output is processed by the referee. The output of the referee is in turn processed by the reverse network which is supposed to reconstruct the input. The network outputs are involved in different loss terms. The few ground-truth masks $\mathbf{y}$ available are used by loss terms relying on supervision \modif{(corresponding data flow in dotted lines)}. Synthetic masks $\mathbf{m}$ are used by the referee network only \modif{(corresponding data flow in gray lines)}. \modif{Real data with or without annotations flow to the consistency and reconstruction loss terms (corresponding data flows in black lines).} \label{fig:pipe}}
\end{figure*}

These four phases involve different loss terms which will be introduced in the next subsections. A global picture of the network interactions with each other and loss terms is shown in Figure \ref{fig:pipe}.


\subsection{Referee training} 
\label{sub:referee_training}

The efficiency of the proposed learning scheme heavily relies on the quality of the trained referee. In order to obtain a referee with strong generalization performances, we propose to leverage the simple structure of region mask images. Indeed, the objects to segment are micro-capsules therefore we know that binary masks have simple elliptical or polygonal shapes\footnote{This is also true for a larger class of biomedical segmentation problems. In particular, many cells can be accurately characterized by elliptical/polygonal masks.}. Such shapes can be easily generated to create a set of synthetic masks. 
In this paper, we randomly generate only ellipses by uniformly sampling the center position, orientation and axis lengths. For axis lengths, the uniform distribution support is $\left[ 0.03\times d;0.6\times d \right] $ where $d^2$ is the number of pixels inside an image. To obtain slightly more general shapes, we apply a random non-linear distortion\footnote{We use the \texttt{random\_distortion} function from the \href{https://augmentor.readthedocs.io/en/master/code.html}{Augmentor} python package with default parameters.} to elliptical masks.

This generating process allows to obtain a very large set of "correct" region masks denoted by $\mathbf{m}$ 
which are targets for the referee. 
For a given flawless $\mathbf{m}$, we can use DA mechanisms to obtain "imperfect" region masks $\tilde{\mathbf{m}}$ by corrupting $\mathbf{m}$. Among other possibilities, we will use a core removal operation followed by standard Gaussian additive noise injection. Core removal consists in the elementwise multiplication of the deformed ellipse mask with its inner contour whose thickness is uniformly drawn between minimal and maximal values. 
Each imperfect/correct pair of region masks is a training example for the referee. The set of these synthetic training data is denoted by $\mathcal{M}$. Given this dataset, the first phase of the learning scenario consists in solving

\begin{equation}\label{eq:loss_synth}
     \underset{\boldsymbol \phi}{\arg\min} \; L_{\textrm{synth}} \left( \boldsymbol \phi ; \mathcal{M}\right) = \sum_{\left( \tilde{\mathbf{m}},\mathbf{m} \right) \in \mathcal{M}} \left\lVert \mathbf{m} - f_{\boldsymbol\phi}^{(\text{ref})}\left(   \tilde{\mathbf{m}} \right)   \right\rVert_2^2,
 \end{equation} 
 which is a mean square error (MSE) loss. An approximate solution to this problem will be obtained by a gradient descent optimizer traditionally used to train neural networks.


\subsection{Trainee pre-training} 
\label{sub:trainee_pre_training}

The second move of the proposed training protocol consists in pre-training the trainee neural network $f_{\boldsymbol\theta}^{(\text{tne})}$. 
To that end, we use the small supervised set $\mathcal{S}_{\textrm{tr}}$ and its validation companion $\mathcal{S}_{\textrm{val}}$. The training set is used to optimize the trainee model as part of the following MSE problem:

\begin{equation}\label{eq:loss_sup}
    \underset{\boldsymbol \theta}{\arg\min} \; L_{\text{sup}}\left( \boldsymbol\theta ; \mathcal{S}_{\textrm{tr}}\right) = \sum_{\left( \mathbf{x},\mathbf{y} \right) \in \mathcal{S}_{\textrm{tr}}} \left\lVert \mathbf{y} - f_{\boldsymbol\theta}^{(\text{tne})}\left( \mathbf{x} \right)   \right\rVert_2^2.
\end{equation}
The validation set $\mathcal{S}_{\textrm{val}}$ is a smaller held-out fraction of the data used for early stopping (ES) in order to detect convergence. As part of ES, the validation data is used to monitor the loss $L_{\text{sup}}\left( \boldsymbol\theta ; \mathcal{S}_{\textrm{val}} \right)$ and when no improvement is observed after a predefined number of "patience" epochs, the training stops. 

\subsection{Reverse pipeline network} 
\label{sub:reverse_pipeline_network}

Similarly as in the previous step, we will again use $\mathcal{S}_{\textrm{tr}}$ to train the reverse net $f_{\boldsymbol \varphi}^{(\text{rev})}$. The main difference is that the roles of inputs and outputs are swapped in the corresponding MSE problem:

\begin{equation}\label{eq:loss_rev}
    \underset{\boldsymbol \varphi}{\arg\min} \; L_{\text{sup}}^{(-1)}\left( \boldsymbol\varphi  ; \mathcal{S}_{\textrm{tr}} \right) = \sum_{\left( \mathbf{x},\mathbf{y} \right) \in \mathcal{S}_{\textrm{tr}}} \left\lVert \mathbf{x} - f_{\boldsymbol\varphi}^{(\text{rev})}\left( \mathbf{y} \right)   \right\rVert_2^2.
\end{equation}
Note that this mask-to-input problem is more difficult than the input-to-mask problem. Training will fail if the distribution of inputs is too complicated. In our case, we exploit the knowledge that micro-capsule 
input images have a rather simple structure and that there is no style variation, i.e. all images have the same texture and illumination features for both background and object regions respectively. Again, this is also true for other several cell segmentation datasets. 
This training is meant to be permanent and therefore has to be successful for the final training phase to work. ES is used in the same way as for the trainee pre-training based on $L_{\text{sup}}^{(-1)}\left( \boldsymbol\theta ; \mathcal{S}_{\textrm{val}} \right)$ values achieved after each epoch.


\subsection{Main training phase} 
\label{sub:train_main}

Once the three previous steps are completed, then a more intensive and final learning episode can start. In this phase, we will continue to optimize $f_{\boldsymbol\theta}^{(\text{tne})}$ but we will keep the referee $f_{\boldsymbol\phi}^{(\text{ref})}$ and $f_{\boldsymbol \varphi}^{(\text{rev})}$ unchanged. We now use all three networks together and define two additional loss terms that exploit unlabeled data $\mathcal{U}_{\textrm{tr}}$ as follows:

\begin{equation}
L_{\textrm{cons}}\left( \boldsymbol \theta ; \mathcal{U}_{\textrm{tr}}\right) = \sum_{\mathbf{x} \in \mathcal{U}_{\textrm{tr}}} \left\lVert f_{\boldsymbol\phi}^{(\text{ref})} \left( f_{\boldsymbol\theta}^{(\text{tne})}\left( \mathbf{x} \right)\right) - f_{\boldsymbol\theta}^{(\text{tne})}\left( \mathbf{x} \right)   \right\rVert_2^2,\label{eq:loss_cons}
\end{equation}
\begin{equation}
L_{\textrm{ae}} \left( \boldsymbol \theta ; \mathcal{U}_{\textrm{tr}}\right) = \sum_{\mathbf{x} \in \mathcal{U}_{\textrm{tr}}} \left\lVert f_{\boldsymbol \varphi}^{(\text{rev})} \left( f_{\boldsymbol\phi}^{(\text{ref})} \left( f_{\boldsymbol\theta}^{(\text{tne})}\left( \mathbf{x} \right)\right) \right) - \mathbf{x}    \right\rVert_2^2.\label{eq:loss_ae}
\end{equation}

$L_{\textrm{cons}}$ is a consistency loss term that forces the trainee to produce region masks that do not deserve any rectification from the referee network. $L_{\textrm{ae}}$ is a reconstruction loss that forces all three networks to work consistently. In particular, it helps preventing from pathological situations where the trainee could converge to a trivial constant black function which the teacher would have no choice but acknowledge as a very good prediction. In the end, in this final training episode, we address the following problem:
\begin{equation}
    \underset{\boldsymbol \theta}{\arg\min} \; L_{\text{sup}}\left( \boldsymbol\theta ; \mathcal{S}_{\textrm{tr}} \right) + L_{\textrm{cons}}\left( \boldsymbol \theta ; \mathcal{U}_{\textrm{tr}} \right) + \lambda_{\textrm{ae}} L_{\textrm{ae}} \left( \boldsymbol \theta ; \mathcal{U}_{\textrm{tr}} \right),\label{eq:loss_global}
\end{equation}
where $\lambda_{\textrm{ae}}$ is a hyperparameter allowing to tune the weight given to the AE loss. Although a similar hyperparameter could be introduced to tune the weight of the consistency loss, it proved unnecessary in our experiments.
\modif{Depending on the amount of annotated data at hand, $\lambda_{\textrm{ae}}$ needs to be adjusted in order to avoid converging to trivial predictors such as an all-black-image predictor.} 
The loss term from the second phase of the protocol is preserved in order to avoid catastrophic forgetting of the supervision brought by $\mathcal{S}_{\textrm{tr}}$. The minimization problem \eqref{eq:loss_global} is stopped using ES which relies on both $\mathcal{S}_{\textrm{val}}$ and $\mathcal{U}_{\textrm{val}}$.

This learning phase is more challenging than the previous ones and it has proved much beneficial to execute it as part of a curriculum learning (CL) procedure \cite{bengio2009curriculum}. CL consists in selecting a subset of data points to start training a model and gradually unveiling the rest after the model has started to converge. The motivation behind CL is that not all training examples are equally easy to learn from and asking to optimize losses w.r.t. the whole dataset might be too much to ask. 
In this paper, CL is exclusively applies to the set $\mathcal{U}_{\textrm{tr}}$. 

To implement CL, one has to choose a selection criteria $s \left( \mathbf{x} \right)$ that tells us how useful is example $\mathbf{x}$ to help the model to start learning efficiently. We thus introduce the following criterion that uses input images only:

\begin{equation}
    s  \left( \mathbf{x} \right) = - \frac{ \left\lVert f_{\boldsymbol \varphi}^{(\text{rev})} \left( f_{\boldsymbol\phi}^{(\text{ref})} \left( f_{\boldsymbol\theta}^{(\text{tne})}\left( \mathbf{x} \right)\right) \right) - \mathbf{x}    \right\rVert_1}{\left\lVert \mathbf{x}\right\rVert_1}.
\end{equation}
The numerator of $s \left( \mathbf{x} \right)$ is meant to promote images for which all three networks work well and the denominator is meant to promote images that contain a rather large object. Note that for the denominator to carry the intended meaning, the object to segment must contain in average brighter pixels than the background does. If it is not the case, one can just pre-process images by mapping $\mathbf{x}$ to $1-\mathbf{x}$. Because the three network are either pre-trained or fully trained, one can readily use $s \left( \mathbf{x} \right)$ before launching the fourth phase of the training protocol. In this paper, we use CL to select $30\%$ of the images contained in $\mathcal{U}_{\textrm{tr}}$ to solve \eqref{eq:loss_global}. We then learn from this subset of images until convergence. The criterion is used again to increase the dataset by $7\%$. We repeat this procedure 10 times and thus finish with a training run using the entire subset $\mathcal{U}_{\textrm{tr}}$. 

Across this CL procedure, we also keep track of the following validation loss 
\begin{equation}
L_{\textrm{val}} = L_{\text{sup}}\left( \boldsymbol\theta ; \mathcal{S}_{\textrm{val}} \right) + L_{\textrm{cons}} \left( \boldsymbol \theta ; \mathcal{U}_{\textrm{val}} \right) + L_{\textrm{ae}} \left( \boldsymbol \theta ; \mathcal{U}_{\textrm{val}}\right).
\end{equation}

We return the trainee network that achieves maximal validation loss. Indeed, the best model is not always the one obtained from the last CL iteration. 
Note that $L_{\textrm{val}}$ re-uses the held-out validation sets used for ES which is not ideal. 
The reasons behind this choice is that $\mathcal{S}_{\textrm{tr}}$ is supposed to be really small in our setting and sacrificing even more images for a second validation set is problematic. 
In addition, the model selection intended behind the computation of $L_{\textrm{val}}$ has little chance to be influenced by the result of ES.


\section{Results}
\label{sec:exp}

In this section, we show that the self-supervised learning scenario proposed in this paper, and called self-mentoring, allows to achieve very good segmentation performances from only a very few labeled images and a moderately large set of unlabeled images. 
A first experiment is conducted on the Capsule dataset where a transparent object (micro-capsule) must be segmented. 
The second subsection of this section also includes experiments where a covariate shift between train and test samples is created in order to assess the robustness of self-mentoring. 
Thirdly, the coupling of self-mentoring with DA is also investigated. 
Finally, although this paper is focused on micro-capsule segmentation, an experiment on non-transparent objects (cells) is also proposed to confirm the validity of the scenario. This experiment re-uses the referee trained in the previous experiments which shows a high level of transferability of the latter. 

Segmentation accuracy is given in terms of average Jaccard index which, for a test sample $\left( \mathbf{x}, \mathbf{y} \right) $ writes

\begin{equation}
        \textrm{JI}(\%) = 100\times \frac{ \left\lVert \mathbf{y} \odot \hat{\mathbf{y}} \right\rVert_1  }{\left\lVert \mathbf{y} + \hat{\mathbf{y}} \right\rVert_1 - \left\lVert \mathbf{y} \odot \hat{\mathbf{y}} \right\rVert_1  }, 
\end{equation}
where the predicted mask $\hat{\mathbf{y}}$ is obtained by comparing $f_{\boldsymbol\theta}(\mathbf{x})$ to a threshold of $0.5$ and $\odot$ is the elementwise product between predicted and ground truth (GT) masks. The numerator of JI is the number of pixels inside the intersection of the masks while the denominator is the number of pixels inside the union of the masks. 
The average is computed over test samples and over 5 different random parameter initializations of the neural networks. 

All networks are U-net instances ($M=3$ for all, $F=5$ for trainee and reverse networks, $F=30$ for the referee) optimized using RMSprop with a learning rate of $1e-4$ and discount factor of $0.9$. 
\modif{The very small number of available annotated images in the targeted few shot learning setting accounts for the hyperparamter choices for U-net instances for the trainee and reverse nets. Higher values for $M$ or $F$ would lead to overfitting during phases 2 and 3 of the the protocol. Conversely, the referee network is trained on synthetic data. Consequently, a network with much larger learning capacity can be learnt since one can sample additional data to avoid overfitting. For values of $F$ above 30, no significant performance gain were observed while requiring more training epochs to converge.}
Although the learning rate could be tailored for each training phase separately, in our experiments, it appeared to be unnecessary. We use a batchsize of one.



\begin{figure}
     \centering
     \begin{subfigure}[t]{0.15\textwidth}
         \centering
         \includegraphics[width=\textwidth]{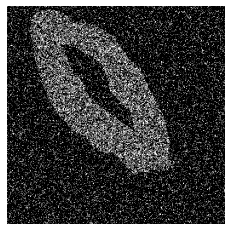}
         \caption{$\scriptstyle\tilde{\mathbf{m}}$}
         \label{fig:y equals x}
     \end{subfigure}
     \begin{subfigure}[t]{0.15\textwidth}
         \centering
         \includegraphics[width=\textwidth]{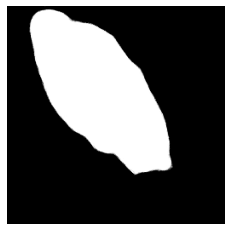}
         \caption{\scalebox{0.8}{$\scriptstyle f_{\boldsymbol\phi}^{(\text{ref})} (\tilde{\mathbf{m}})$}}
         \label{fig:three sin x}
     \end{subfigure}
     \begin{subfigure}[t]{0.15\textwidth}
         \centering
         \includegraphics[width=\textwidth]{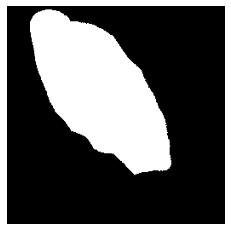}
         \caption{\scalebox{0.8}{$ \mathbf{m}$}}
         \label{fig:five over x}
     \end{subfigure}

        \caption{Example of a synthetic training image pair for the referee as well as the referee output.}
        \label{fig:ref}        
\end{figure}

A very important aspect of the experiments reported in this section is that the same referee network will be used in all of them to assess the strength and the transferability of the latter. For this network, we use a synthetic training dataset $\mathcal{M}$ of size 300 to minimize loss \eqref{eq:loss_synth}. Because synthetic data come for free, we can re-sample a new dataset $\mathcal{M}$. 
This means we actually minimize $\mathbb{E} \left[ L_{\textrm{synth}} \right] $ where the expectation is taken upon the synthetic data distribution. Like for other networks, we also use ES to detect convergence relying on a synthetic validation set of size 300. With a patience parameter of 500, the referee model is obtained in approximately 2500 to 3000 epochs. 
An example of a generated training example for the referee as well as its output is shown in Fig. \ref{fig:ref}

\subsection{Validation of the pipeline on the Capsule dataset} 
\label{sub:validation_of_the_pipeline_on_the_capsule_dataset}


A held-out test set of 50 images from the Capsule dataset is kept to compute Jaccard indices assessing the performances of the tested approaches. 
To place ourselves in a challenging semi-supervised FSL setting for which our self-supervised pipeline was designed, only $n_{\textrm{s-tr}} \in \left\{2; 3;4;5;6 \right\}$ + $n_{\textrm{s-val}}=1$ labeled images will be actually used to train U-net. 
For the unlabeled data, we will use $n_{\textrm{u-val}} =20$ images in the unsupervised validation set and $n_{\textrm{u-tr}} = 281 $ for the unsupervised training set (dropping the corresponding labels). 

Since the first phase of the self-mentoring pipeline (referee training) has already been carried out, we now proceed to the remaining steps.  
For the second and third phases of the learning protocol, the training is much faster due to the very limited amount of data in $\mathcal{S}_{\textrm{tr}}$. A patience of 20 is used for ES with validation data $\mathcal{S}_{\textrm{val}}$ for both of them. 
In the final and fourth phases, only the trainee is further trained. Inside each step of the CL loop described in \ref{sub:train_main}, we use again ES with a patience of 40. 

This subsection includes a comparison with self-training \cite{xie2020self} in which a teacher is first trained on $\mathcal{S}_{\textrm{tr}}$ with ES on $\mathcal{S}_{\textrm{val}}$ and is used to create pseudo-labels for $\mathcal{U}_{\textrm{tr}}$ and $\mathcal{U}_{\textrm{val}}$. A student U-net is then retrained to minimize MSE loss involving all training pairs (including the one with pseudo labels). ES is also employed on all validation pairs (also including the one with pseudo labels). 

Segmentation accuracies achieved by U-net in standalone training or through the proposed protocol are reported in Table \ref{tab:acc_caps}. 
For the self-mentoring approach, the reported results correspond to the performances achieved by the selected model (based on validation loss values) after CL is completed. 
As the number of annotated data increase, the results tend to improve for all tested methods. 
By simply cleverly using unsupervised images and leveraging knowledge acquired from synthetic data, self-mentoring exhibits accuracy increments from $+27.09\%$ to $+36.07\%$. 
These performances increments are very significant and prove the validity of our self-mentoring approach in the FSL task addressed in this paper. 
The results obtained by self-training are rather disappointing and this concurrent approach seems not to be able to provide meaningful supervisory signal in this FSL setting. 
Note that the protocol starts struggling to converge when $n_{\textrm{s-tr}}+n_{\textrm{s-val}} \leq 3$. We will see in subsection \ref{sub:coupling_with_data_augmentation}, that DA can come at the rescue in such more extreme FSL settings. 

\begin{figure*}
     \centering
     \begin{subfigure}[t]{0.2\textwidth}
         \centering
         \includegraphics[width=\textwidth]{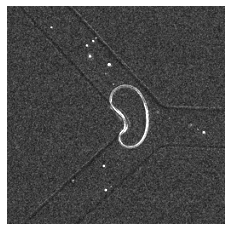}
         \caption{$\scriptstyle\mathbf{x}$}
         \label{fig:y equals x}
     \end{subfigure}
     \begin{subfigure}[t]{0.2\textwidth}
         \centering
         \includegraphics[width=\textwidth]{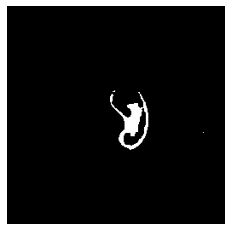}
         \caption{\scalebox{0.8}{$\scriptstyle f_{\boldsymbol\theta}^{(\textrm{tne})}(\mathbf{x})$}}
         \label{fig:three sin x}
     \end{subfigure}
     \begin{subfigure}[t]{0.2\textwidth}
         \centering
         \includegraphics[width=\textwidth]{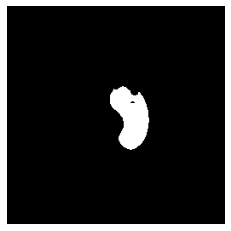}
         \caption{\scalebox{0.8}{$\scriptstyle f_{\boldsymbol\phi}^{(\text{ref})} \left( f_{\boldsymbol\theta}^{(\text{tne})}\left( \mathbf{x} \right)\right)$}}
         \label{fig:five over x}
     \end{subfigure}
     \begin{subfigure}[t]{0.2\textwidth}
         \centering
         \includegraphics[width=\textwidth]{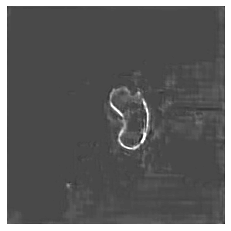}
         \caption{\scalebox{0.8}{$\scriptscriptstyle f_{\boldsymbol \varphi}^{(\text{rev})} \left( f_{\boldsymbol\phi}^{(\text{ref})} \left( f_{\boldsymbol\theta}^{(\text{tne})}\left( \mathbf{x} \right)\right) \right)$}}
         \label{fig:five over x}
     \end{subfigure}
        \caption{Example of an input image for the Capsule dataset along with the corresponding outputs of the 3 networks. These outputs are typically those obtained after the three first phases of the training protocol are done but before the fourth one. After the final phase is performed too, the output $f_{\boldsymbol\theta}^{(\textrm{tne})}(\mathbf{x})$ of the trainee will be much closer to the output $f_{\boldsymbol\phi}^{(\text{ref})} \left( f_{\boldsymbol\theta}^{(\text{tne})}\left( \mathbf{x} \right)\right)$ of the referee.}
        \label{fig:three graphs}        
\end{figure*}

\begin{figure*}
     \centering
     \begin{tabular}{m{1.3cm}c|ccccc}
     & trainee output & \multicolumn{5}{c}{trainee 1st layer feature maps} \\
     before phase 4& \raisebox{-.5\height}{\includegraphics[width=.12\textwidth]{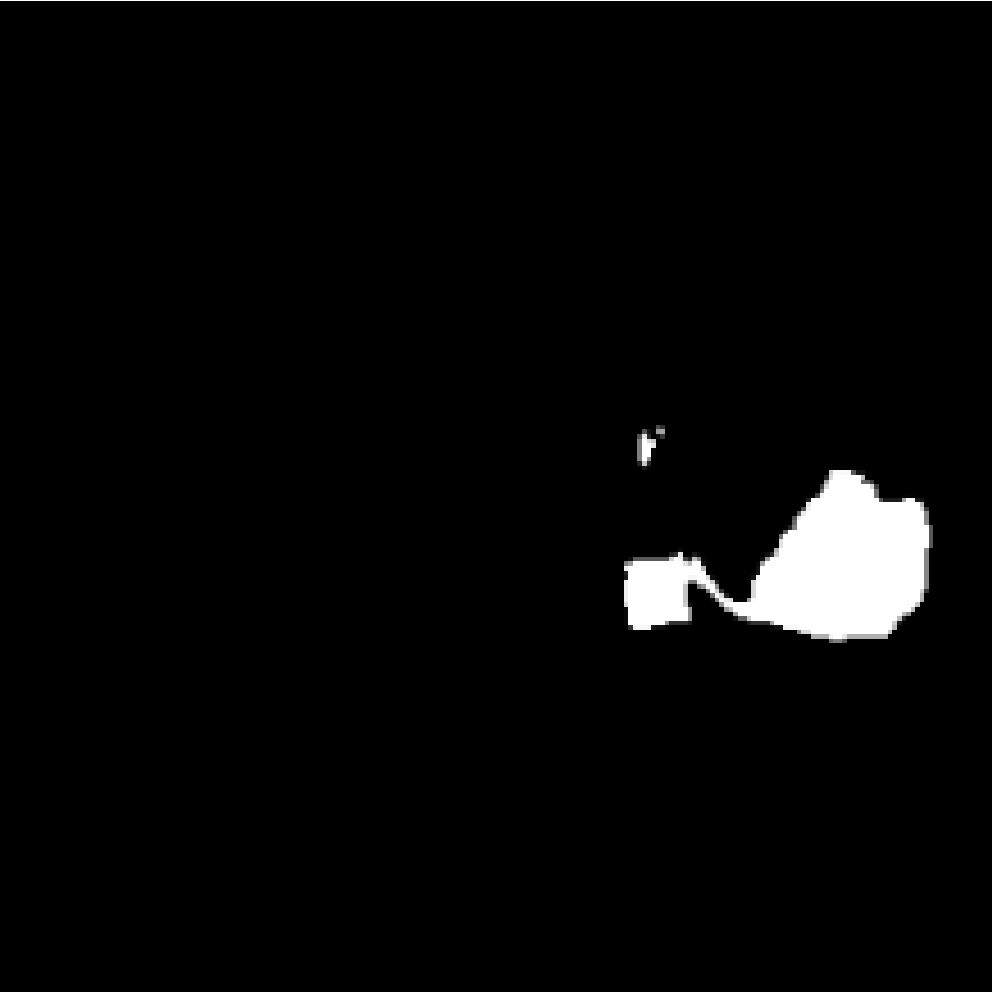}}
     &\raisebox{-.5\height}{\includegraphics[width=.12\textwidth]{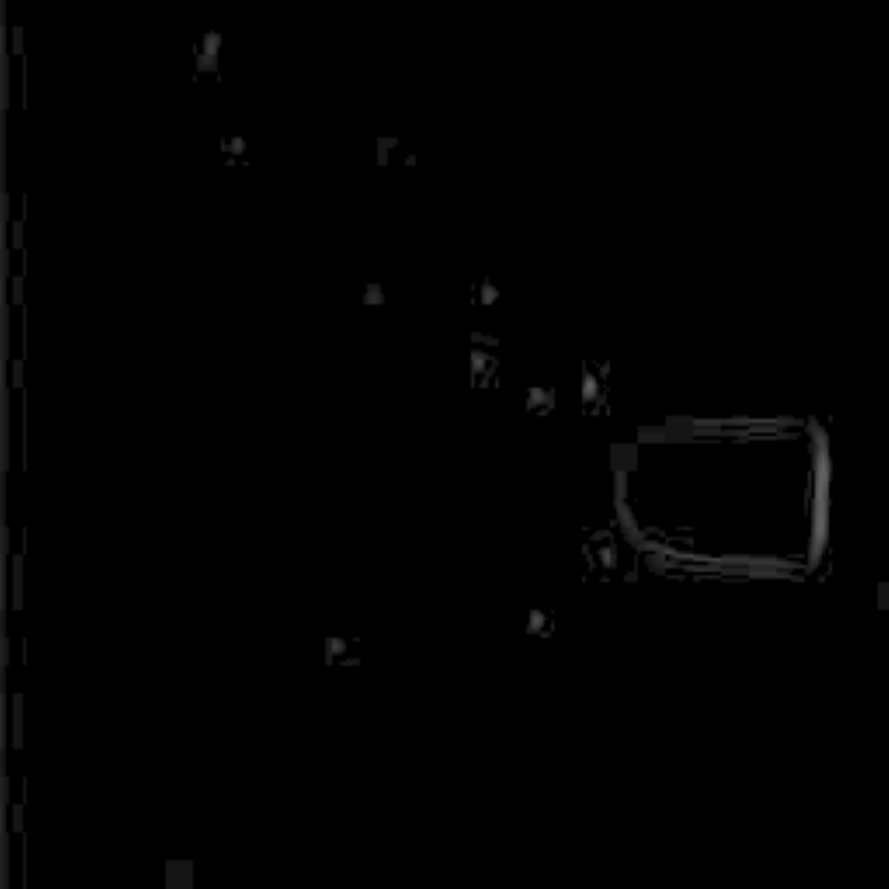}}
     &\raisebox{-.5\height}{\includegraphics[width=.12\textwidth]{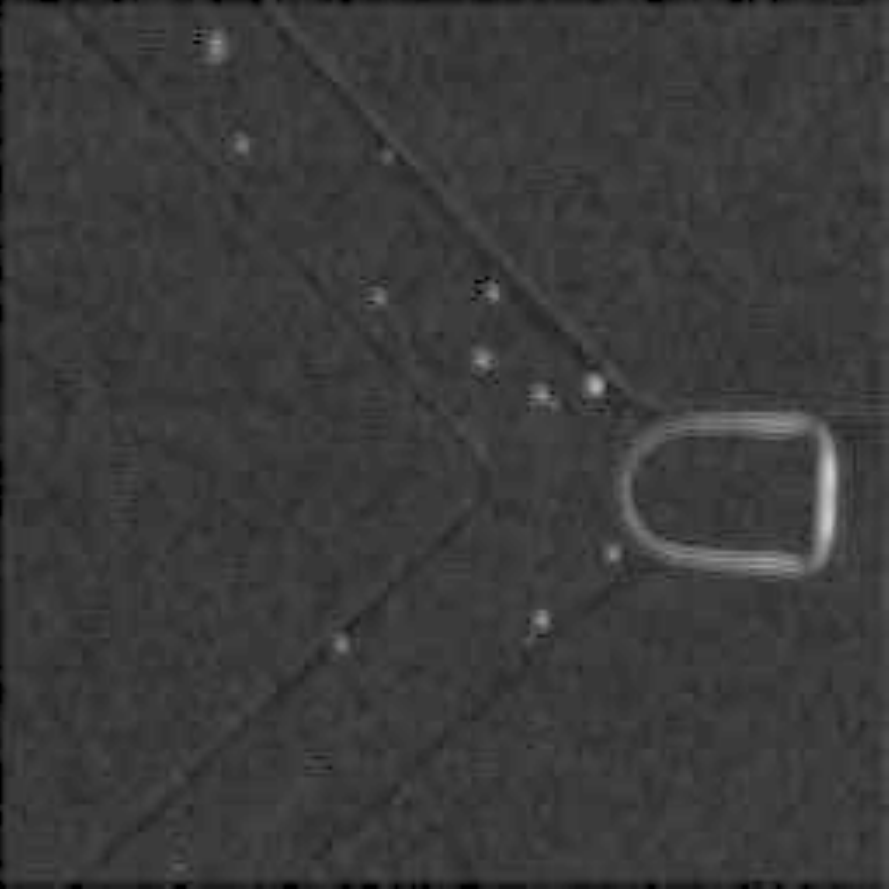}} 
     &\raisebox{-.5\height}{\includegraphics[width=.12\textwidth]{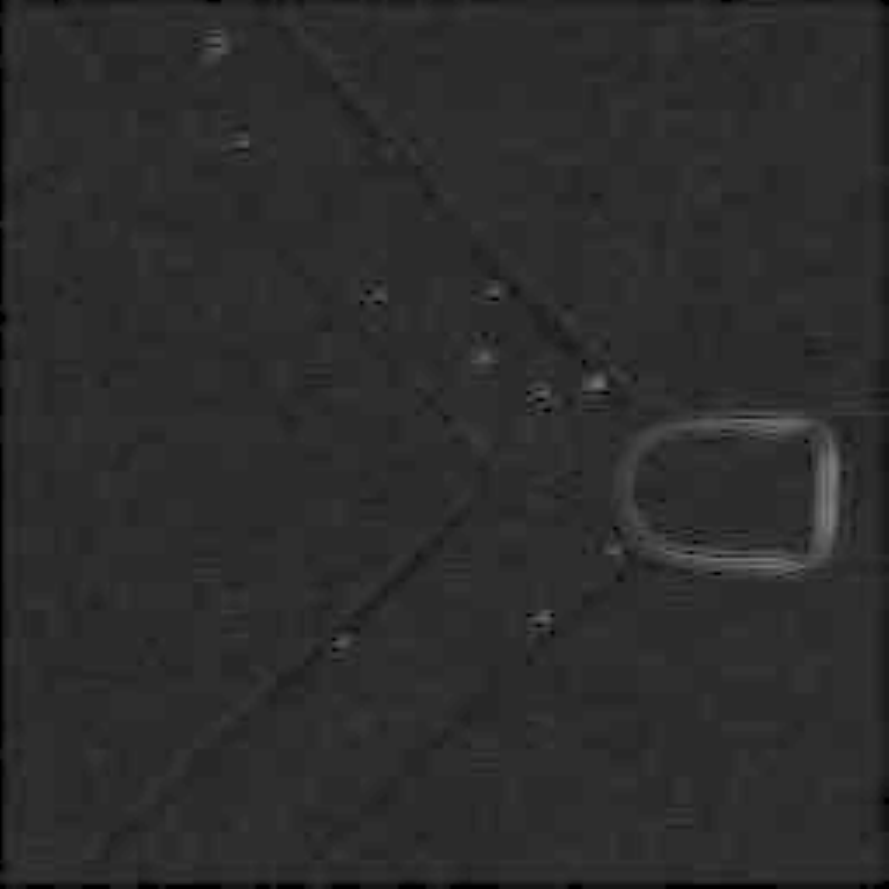}} 
     &\raisebox{-.5\height}{\includegraphics[width=.12\textwidth]{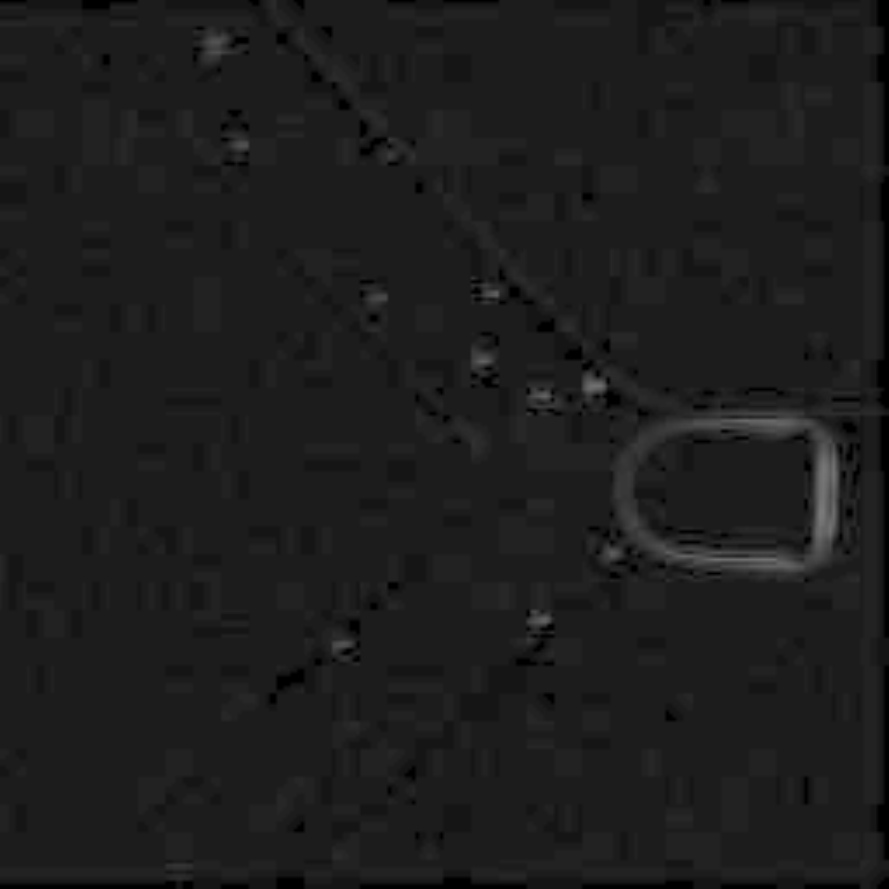}} 
     &\raisebox{-.5\height}{\includegraphics[width=.12\textwidth]{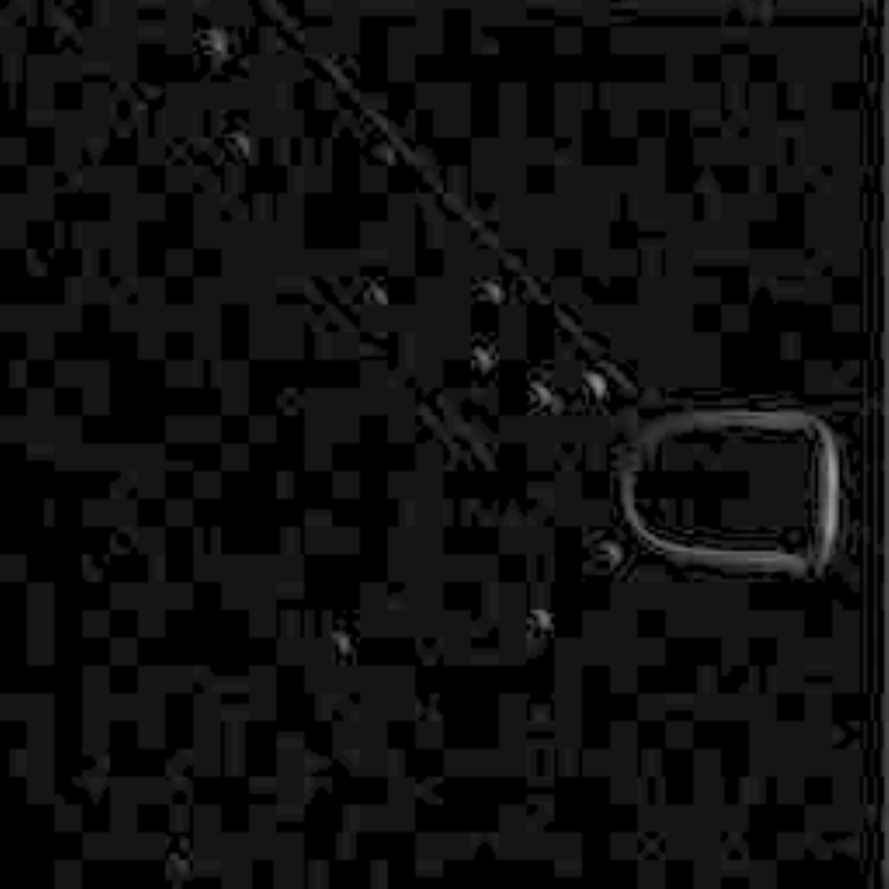}}\\
     
     &&&&&&\\
     
     after phase 4& \raisebox{-.5\height}{\includegraphics[width=.12\textwidth]{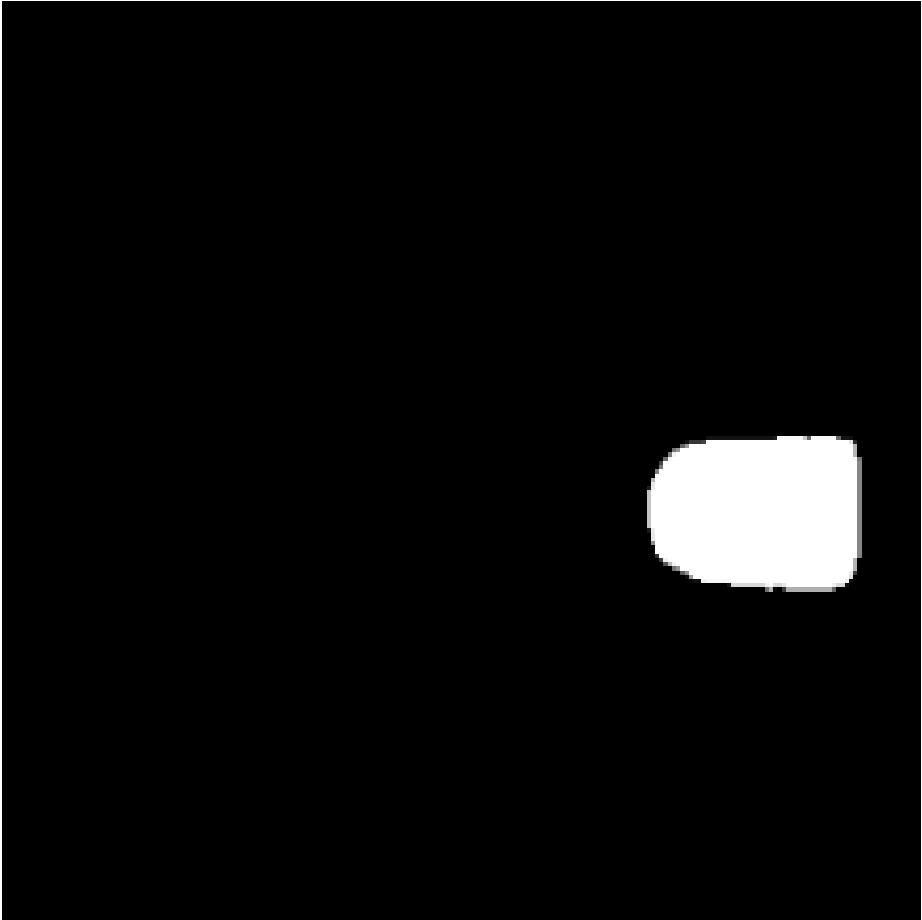}}
     &\raisebox{-.5\height}{\includegraphics[width=.12\textwidth]{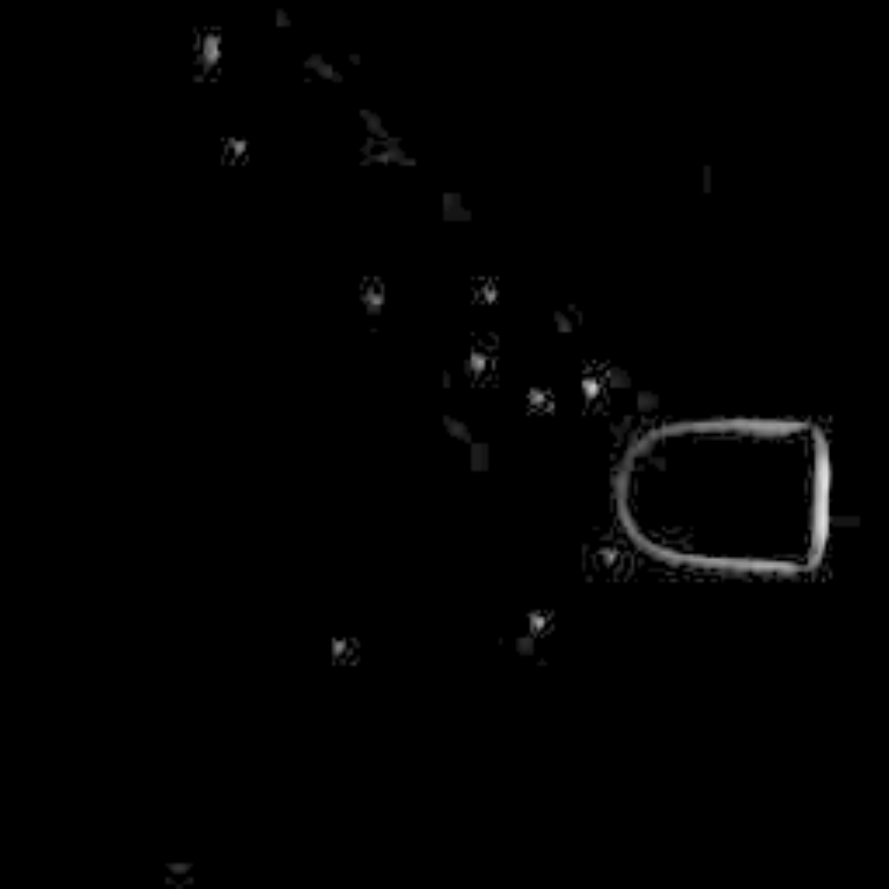}} 
     &\raisebox{-.5\height}{\includegraphics[width=.12\textwidth]{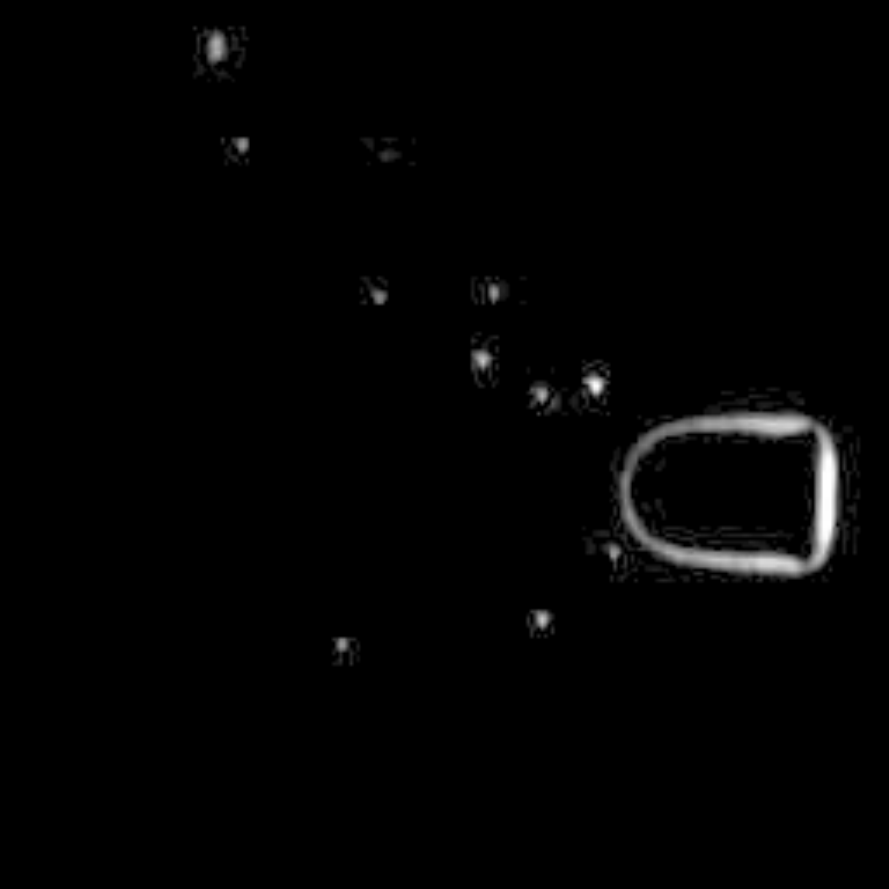}} 
     &\raisebox{-.5\height}{\includegraphics[width=.12\textwidth]{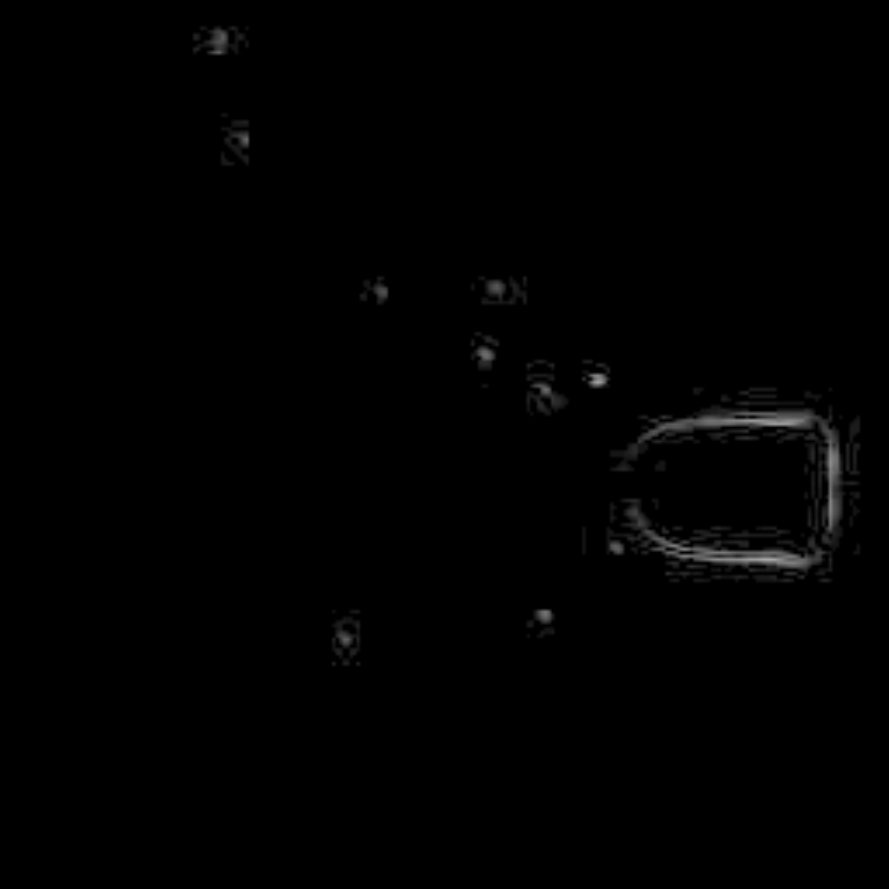}} 
     &\raisebox{-.5\height}{\includegraphics[width=.12\textwidth]{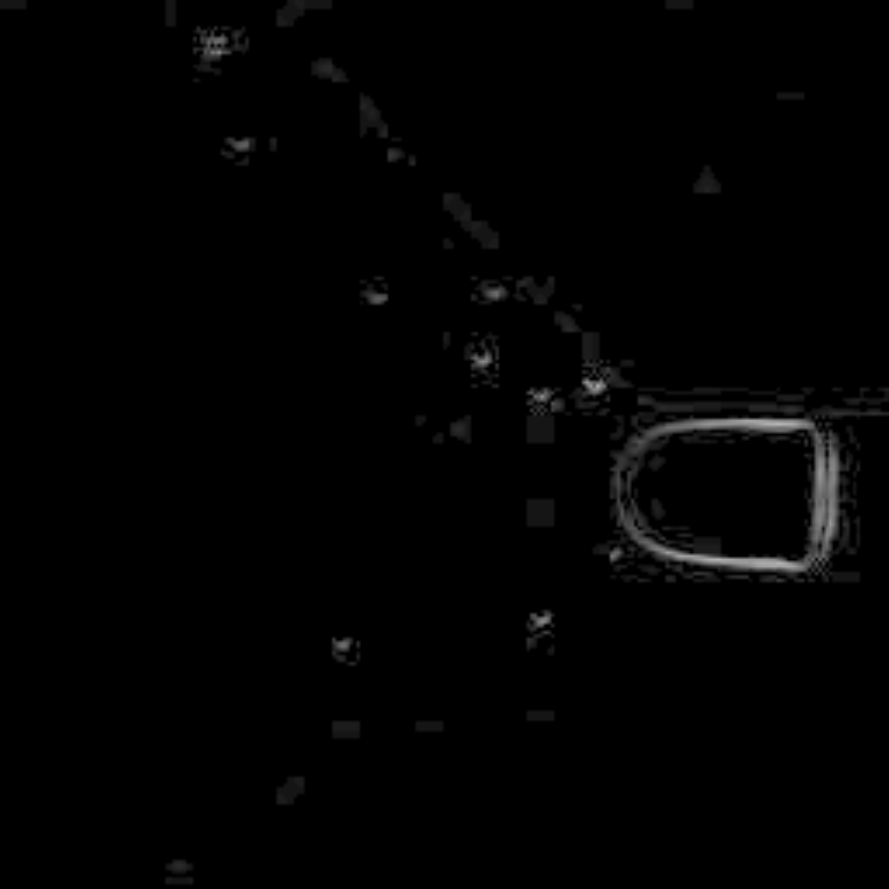}} 
     &\raisebox{-.5\height}{\includegraphics[width=.12\textwidth]{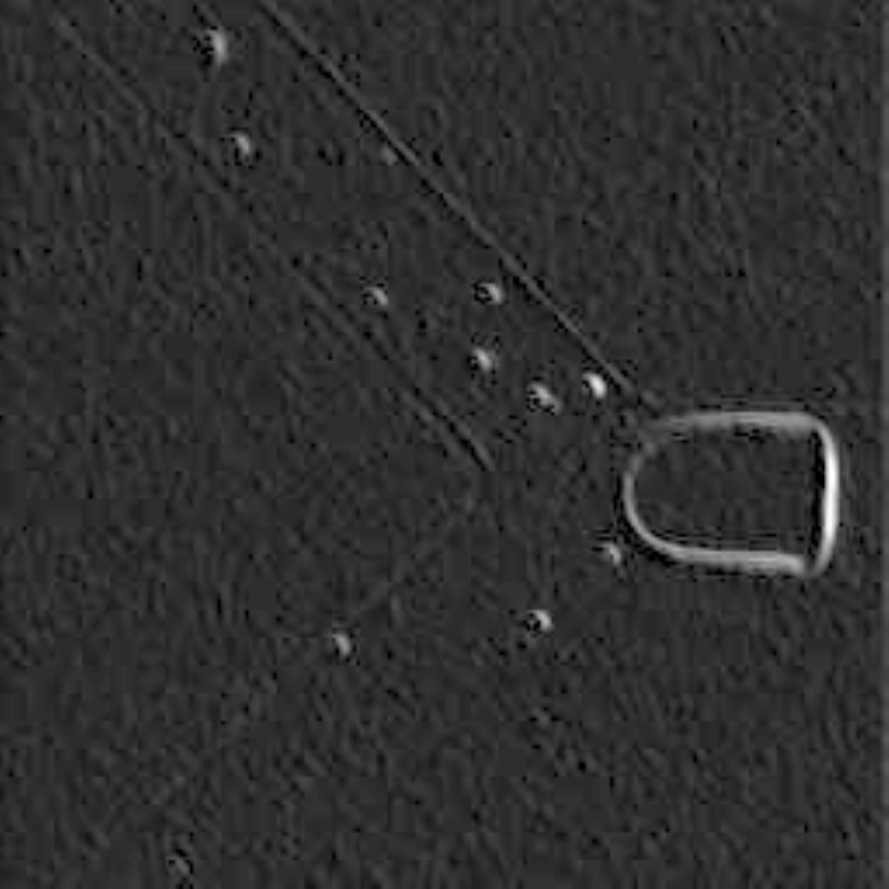}}
     \end{tabular}
     \caption{\modif{Feature maps after the first convolutional layer of the trainee: first row after phase 3 but before phase 4 of the training protocol, second row after the whole training protocol is completed.}\label{fig:first_feat}}
\end{figure*}

\begin{figure*}
     \centering
     \begin{tabular}{m{1.3cm}c|ccccc}
     & trainee output & \multicolumn{5}{c}{trainee 1st layer feature maps} \\
     before phase 4& \raisebox{-.5\height}{\includegraphics[width=.12\textwidth]{pred_li_crop.png}}
     &\raisebox{-.5\height}{\includegraphics[width=.12\textwidth]{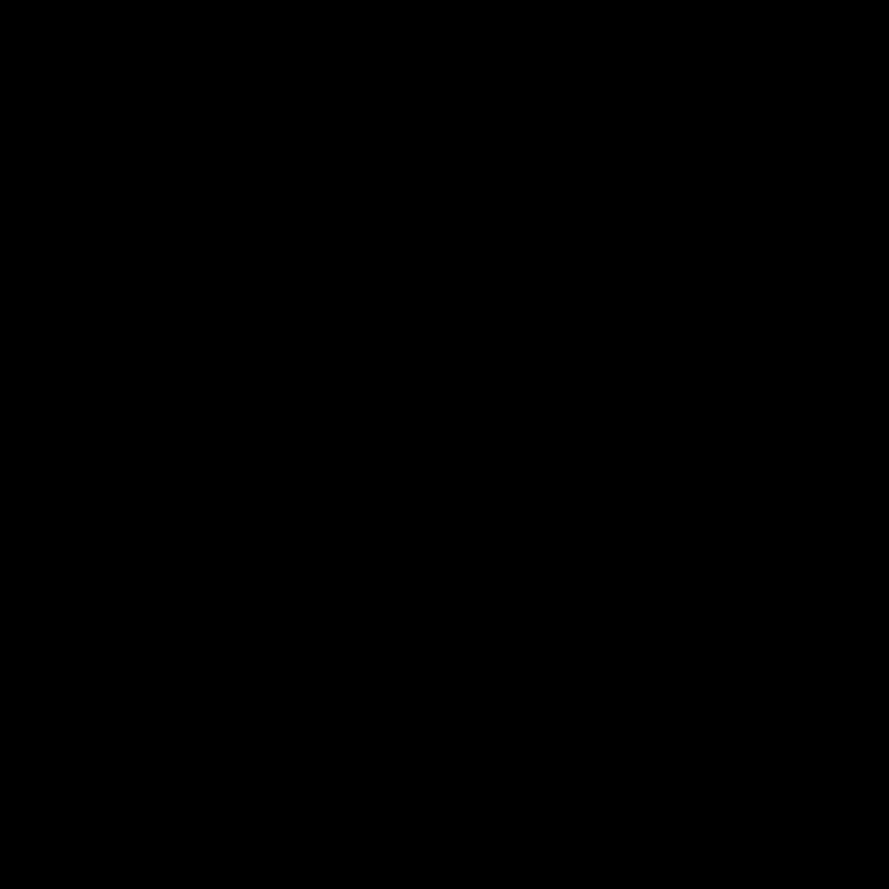}} 
     &\raisebox{-.5\height}{\includegraphics[width=.12\textwidth]{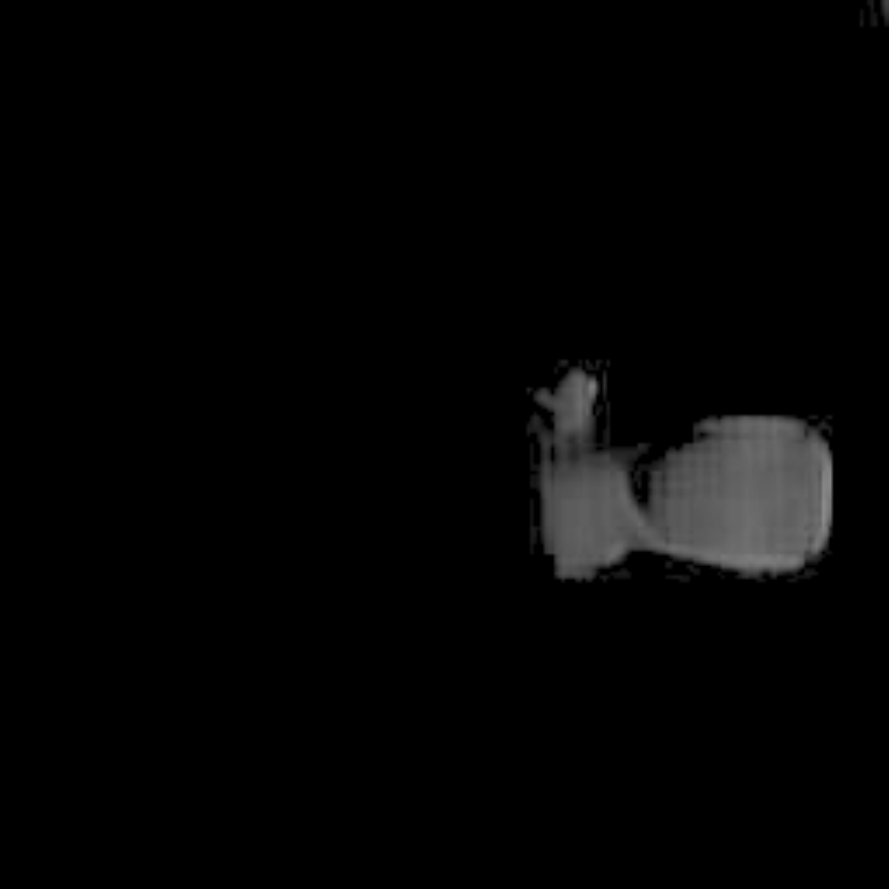}} 
     &\raisebox{-.5\height}{\includegraphics[width=.12\textwidth]{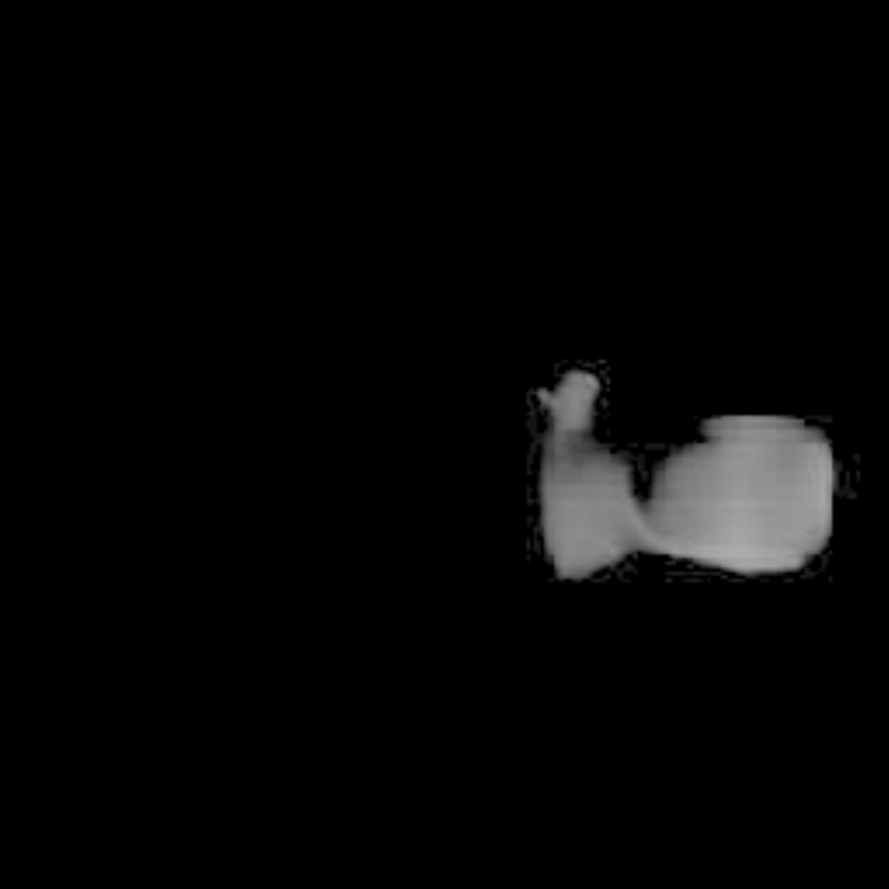}} 
     &\raisebox{-.5\height}{\includegraphics[width=.12\textwidth]{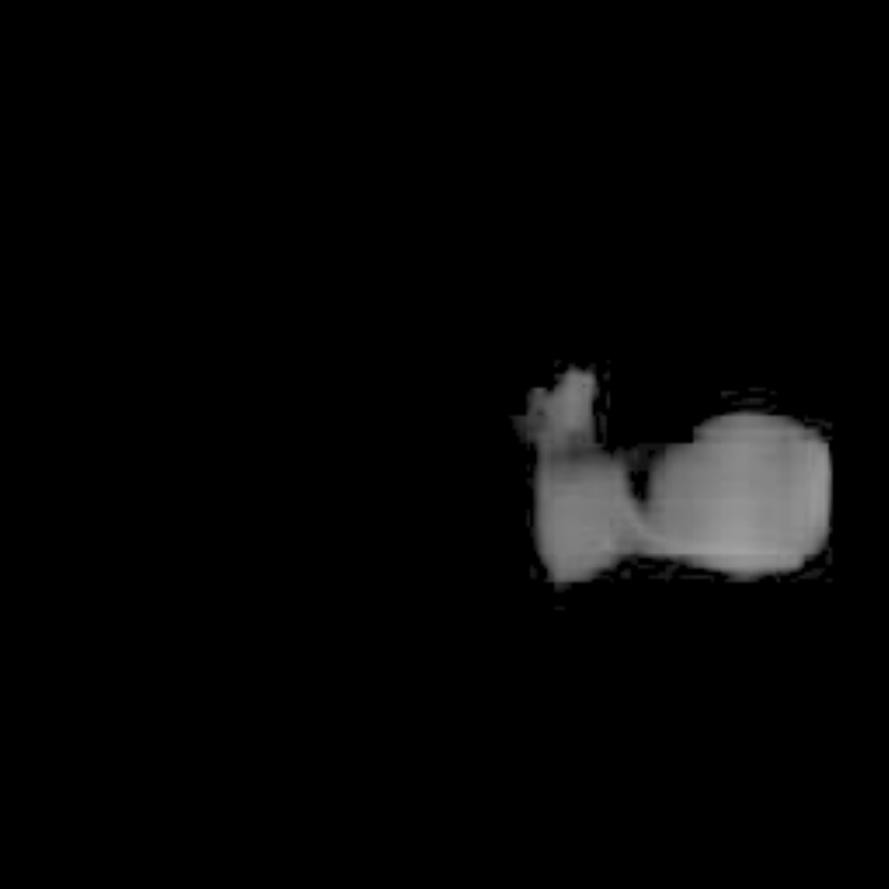}} 
     &\raisebox{-.5\height}{\includegraphics[width=.12\textwidth]{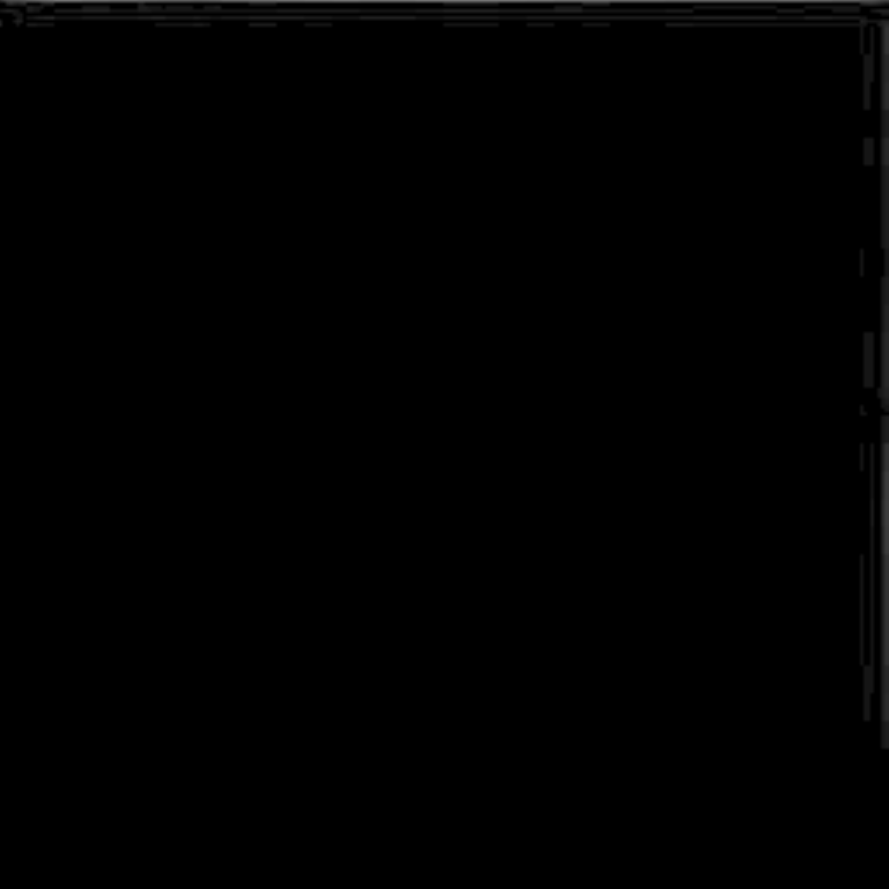}}\\
     
     &&&&&&\\
     
     after phase 4& \raisebox{-.5\height}{\includegraphics[width=.12\textwidth]{pred_lst_crop.png}}
     &\raisebox{-.5\height}{\includegraphics[width=.12\textwidth]{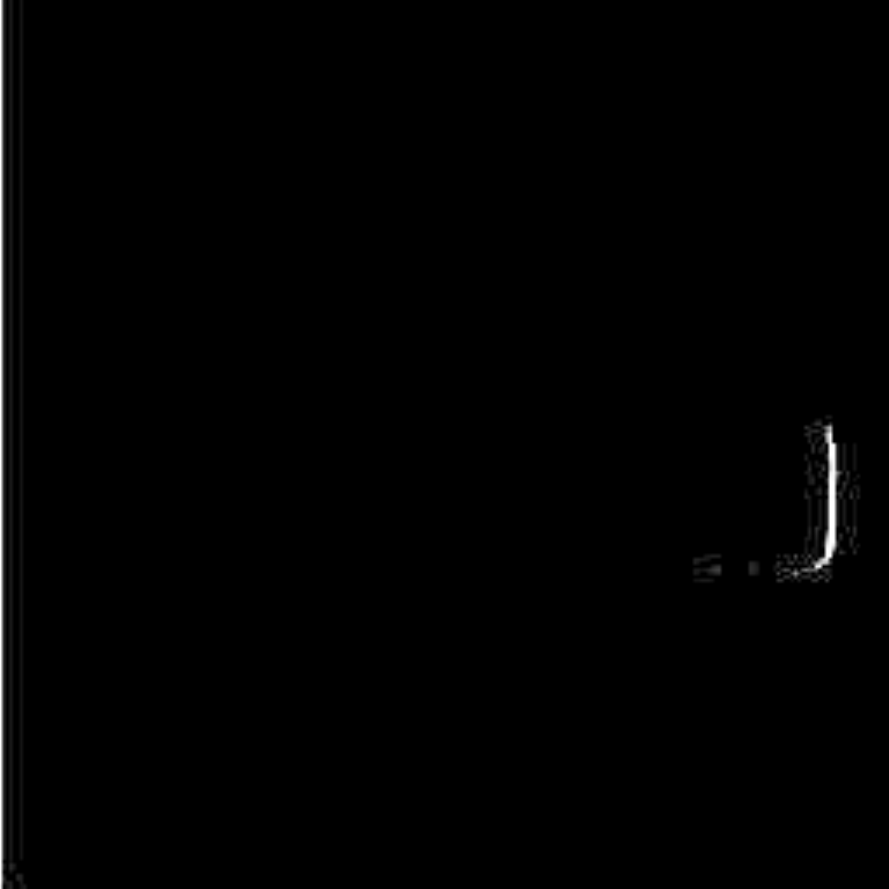}} 
     &\raisebox{-.5\height}{\includegraphics[width=.12\textwidth]{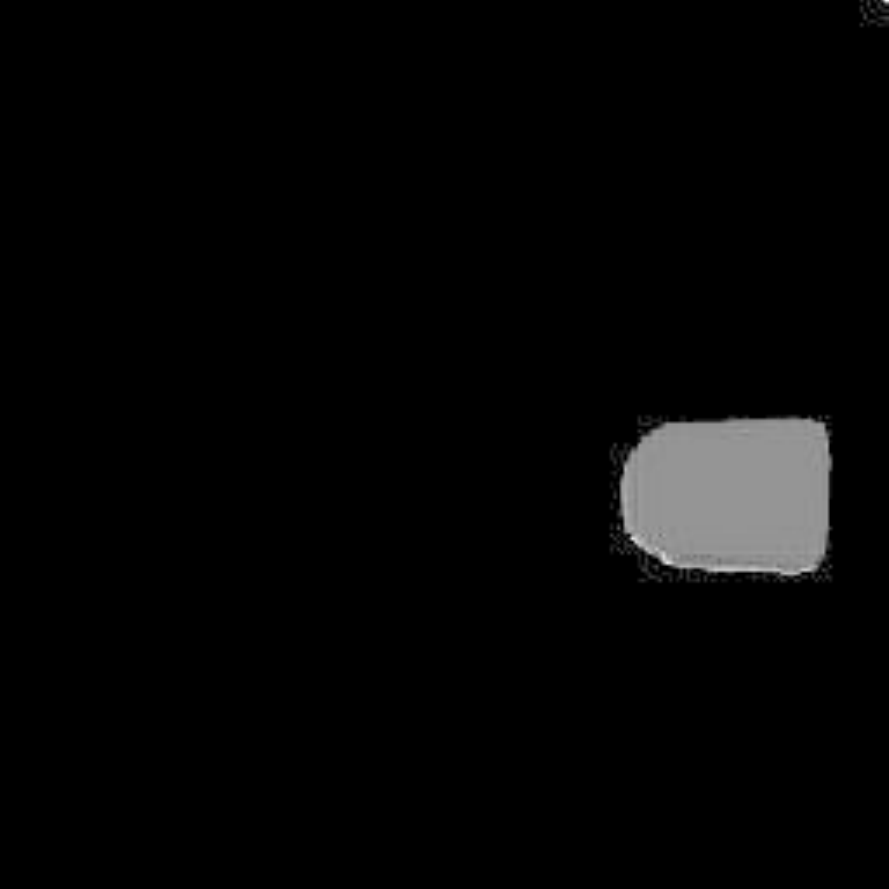}} 
     &\raisebox{-.5\height}{\includegraphics[width=.12\textwidth]{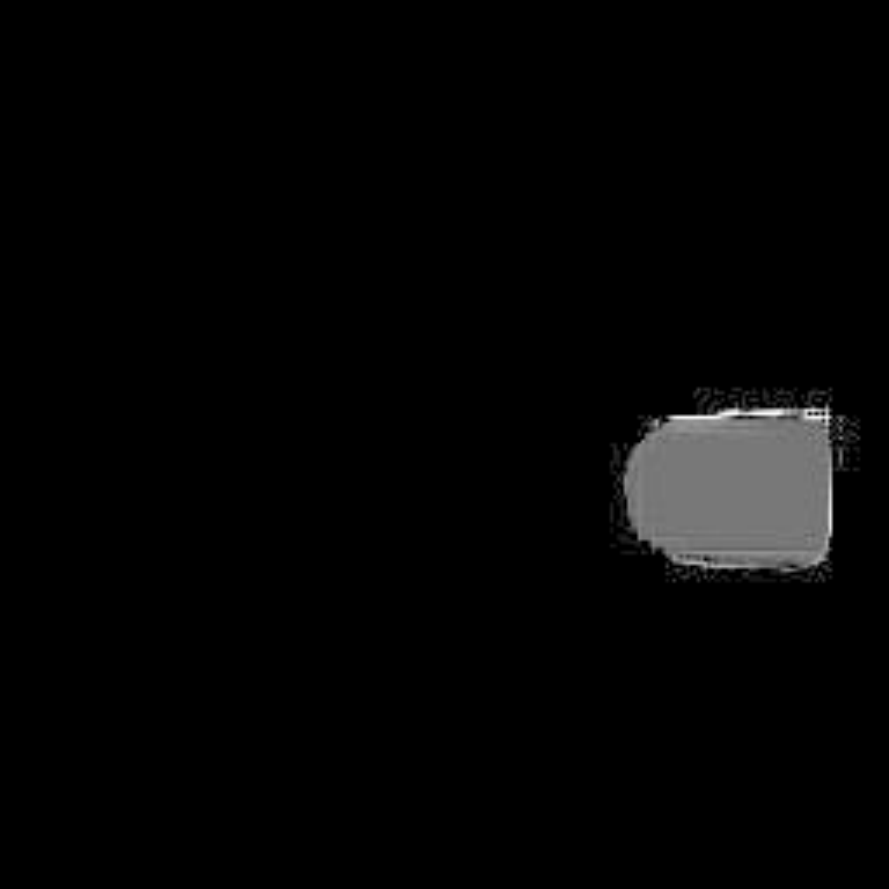}} 
     &\raisebox{-.5\height}{\includegraphics[width=.12\textwidth]{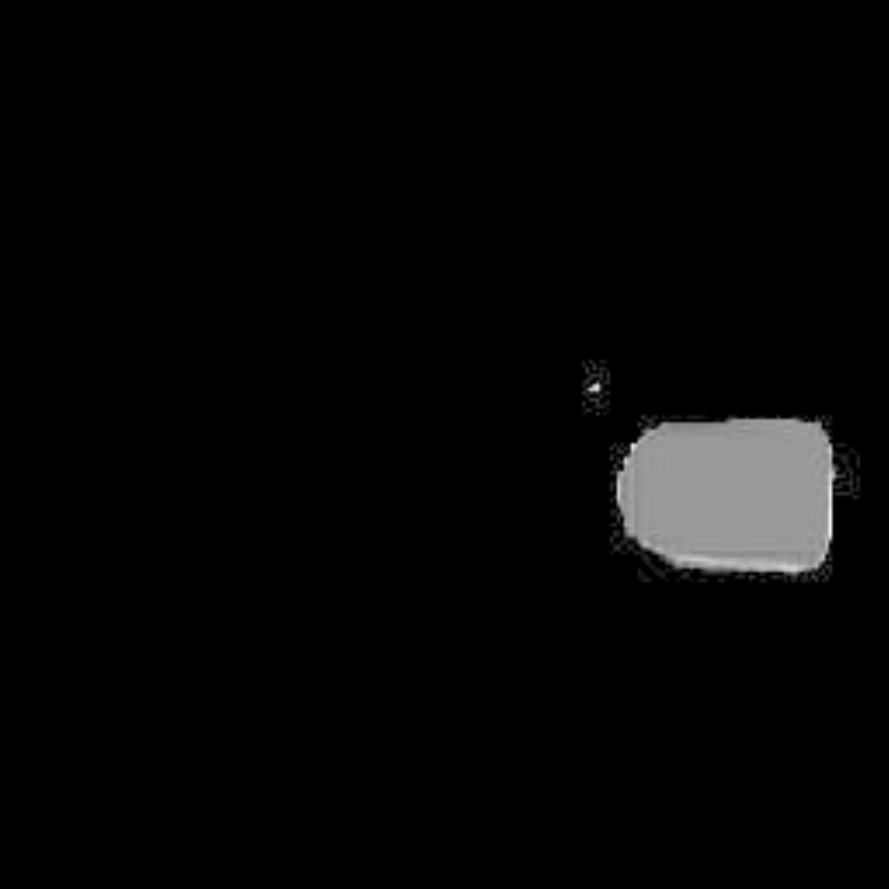}} 
     &\raisebox{-.5\height}{\includegraphics[width=.12\textwidth]{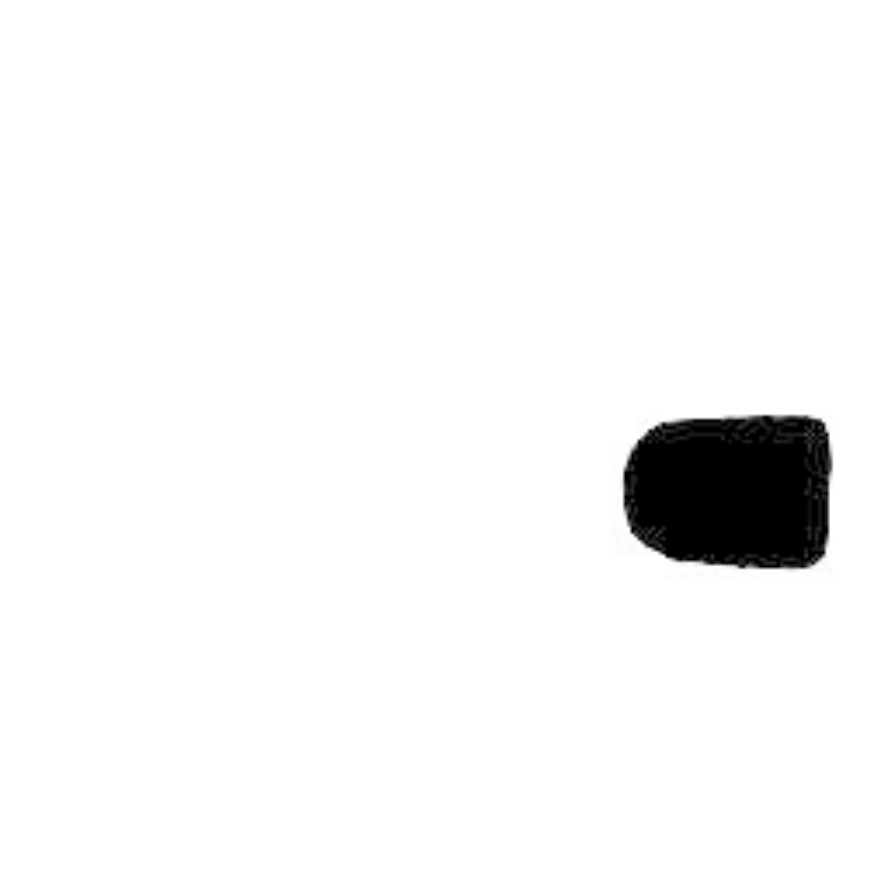}}     
     \end{tabular}
     \caption{\modif{Feature maps after the last convolutional layer of the trainee: first row after phase 3 but before phase 4 of the training protocol, second row after the whole training protocol is completed.}\label{fig:last_feat}}
\end{figure*}

An example of the input/output images through the proposed neural net pipeline are shown in Fig. \ref{fig:three graphs}. 
\modif{To better assess the impact of the knowledge transfer from the referee to the trainee, we also provide feature maps from the first and last layers of the trainee before and after completing the fourth phase of the protocol in Fig.\ref{fig:first_feat} and \ref{fig:last_feat} respectively. We can see that the action of the referee allows the trainee to converge to more acute edge detectors in the first layer and better pre-estimates of the segmentation masks in the last layer. }

\begin{table}[!h]
\begin{center}
\resizebox{\columnwidth}{!}{
 \begin{tabular}{ c c c | c c c }
 \hline
    $n_{\textrm{s-tr}}$& $n_{\textrm{s-val}}$ & $\lambda_{\textrm{ae}}$ & U-net & Self-trained &  Self-mentored\\
     & & & & U-net &  U-net (ours) \\
 \hline 
   2 & 1 & 20 & 27.05\% & 23.63\% & 54.86\% \\
   3 & 1 & 5 & 41.67\% & 41.01\% & 78.64\% \\
   4 & 1 & 100 & 52.67\% & 51.18\% & 80.77\% \\
   5 & 1 & 20 & 56.77\% & 56.09\% & 82.23\% \\
   6 & 1 & 5 & 55.14\% & 55.38\% & 82.43\% \\
   \modif{10} & \modif{1} & \modif{5} & \modif{70.04\%} & \modif{$71.29\%$} & \modif{80.39\%}\\
   \modif{15} & \modif{3} & \modif{1} & \modif{76.95\%} & \modif{$75.50\%$} & \modif{80.28\%}\\
   \modif{25} & \modif{5} & \modif{1}\textbf{} & \modif{80.82\%} & \modif{$81.56\%$} & \modif{85.87\%}\\
   \modif{50} & \modif{10} & \modif{5} & \modif{88.42\%} & \modif{$88.54\%$} & \modif{89.53\%}\\
   \modif{100} & \modif{20} & \modif{1} & \modif{91.42\%} & \modif{$90.59\%$} & \modif{92.29\%}\\
  \hline 
 \end{tabular}}
 \end{center}
 \vspace{-0.5em}
 \caption{Segmentation accuracy (Jaccard Indices) with and without self-supervision on the Capsule dataset. \label{tab:acc_caps}}
\end{table}

\modif{A few complementary results are also provided to assess that the benefits of self-mentoring can also be observed when using different types of loss functions. Indeed, when comparing binary images, it is common to use binary cross-entropy (CE) or the DICE loss. Results when using CE or DICE as part of $L_{\textrm{sup}}$ and $L_{\textrm{cons}}$ are reported in Table \ref{tab:acc_caps_cee} and \ref{tab:acc_caps_dice} respectively. 
It can be observed from Table \ref{tab:acc_caps_cee} that self-mentoring seems to achieve similar performances compared to Table \ref{tab:acc_caps} where only MSE based loss terms are used. However, instabilities during training were observed when using CE loss terms. 
During phase 2 of the protocol, if U-net fails to optimize itself after 5 epochs, the training is stopped and re-started from a different initial vector of parameters. In the case of the DICE loss results reported in Table \ref{tab:acc_caps_dice}, the increment offered by self-mentoring is a bit smaller but still significant. Instabilities were also observed with this loss. These results confirm the benefits of using self-mentoring which is not particularly tied to a given loss choice, although experimental results suggest to use MSE only for a safer training.}

\begin{table}[!h]
\begin{center}
\resizebox{\columnwidth}{!}{
 \begin{tabular}{ c c c | c c c }
 \hline
    $n_{\textrm{s-tr}}$& $n_{\textrm{s-val}}$ & $\lambda_{\textrm{ae}}$ & U-net & Self-trained &  Self-mentored\\
     & & & & U-net &  U-net (ours) \\
 \hline 
   \modif{3} & \modif{1} & \modif{5} & \modif{37.98\%} &  \modif{$36.22\%$} & \modif{$81.40\%^{*}$}\\ 
   \modif{4} & \modif{1} & \modif{5} & \modif{52.24\%} & 
   \modif{$49.56\%$} & \modif{$79.05\%^{*}$}\\    
   \modif{5} & \modif{1} & \modif{5} & \modif{53.58\%} & \modif{$51.29\%$}& \modif{82.19\%}\\

  \hline 
 \end{tabular}}
 \end{center}
 \vspace{-0.5em}
 \caption{\modif{Segmentation accuracy (Jaccard Indices) with and without self-supervision on the Capsule dataset when using Cross-Entropy loss for loss terms $L_{\textrm{sup}}$ and $L_{\textrm{cons}}$. Results of self-mentoring which required re-starting training when a convergence failure is detected in the early epochs are denoted with a $^{*}$ symbol. \label{tab:acc_caps_cee}}}
\end{table}

\begin{table}[!h]
\begin{center}
\resizebox{\columnwidth}{!}{
 \begin{tabular}{ c c c | c c c }
 \hline
    $n_{\textrm{s-tr}}$& $n_{\textrm{s-val}}$ & $\lambda_{\textrm{ae}}$ & U-net & Self-trained &  Self-mentored\\
     & & & & U-net &  U-net (ours) \\
 \hline 
   \modif{3} & \modif{1} & \modif{5} & \modif{41.86\%} &  \modif{$43.10\%$} & \modif{$66.01\%^{*}$}\\ 
   \modif{4} & \modif{1} & \modif{10} & \modif{55.10\%} & 
   \modif{$54.87\%$} & \modif{$77.79\%$}\\    
   \modif{5} & \modif{1} & \modif{20} & \modif{56.59\%} & \modif{$55.12\%$}& \modif{$76.58\%$}\\

  \hline 
 \end{tabular}}
 \end{center}
 \vspace{-0.5em}
 \caption{\modif{Segmentation accuracy (Jaccard Indices) with and without self-supervision on the Capsule dataset when using Dice loss for loss terms $L_{\textrm{sup}}$ and $L_{\textrm{cons}}$. Results of self-mentoring which required re-starting training when a convergence failure is detected in the early epochs are denoted with a $^{*}$ symbol. \label{tab:acc_caps_dice}}}
\end{table}

\modif{In addition, it is also important for fairness of comparison to examine what U-net would be able to do if it were granted the same number of parameters than what all three networks combined use in self-mentoring which is around 8 millions of parameters (192,000 for the trainee). The performances of U-net with equal learning-capacity are reported in Table \ref{tab:acc_caps_large}. It can be observed that, in severe few shot learning settings, self-mentoring consistently improves over a larger U-net by a comfortable margin. However, when moderately few annotated examples are available, the large U-net starts competing with our framework. In this experiment, when 30 annotated images are used, the large U-net outperforms self-mentoring.
Note that, even if the two models have comparable performances, self-mentoring has the advantage of allowing to deploy a much lighter neural network at inference time which is also interesting from a neural network compression standpoint.}

\begin{table}[!h]
\begin{center}
 \begin{tabular}{ c c c | c c }
 \hline
    $n_{\textrm{s-tr}}$& $n_{\textrm{s-val}}$ & $\lambda_{\textrm{ae}}$ & \modif{Large} &  Self-mentored\\
     & & & \modif{U-net}  &  U-net (ours) \\
 \hline 
   2 & 1 & 20 & \modif{34.72\%}  & 54.86\%\\
   3 & 1 & 5 & \modif{51.20\%}  & 78.64\%\\
   4 & 1 & 100 & \modif{65.90\%}  & 80.77\%\\
   5 & 1 & 20 & \modif{65.94\%}  & 82.23\%\\
   6 & 1 & 5 & \modif{69.29\%}  & 82.43\%\\
   25 & 5 & 1 & \modif{87.17\%}  & 85.87\%\\
  \hline 
 \end{tabular}
 \end{center}
 \vspace{-0.5em}
 \caption{\modif{Equal learning-capacity segmentation accuracy (Jaccard Indices) comparison. The results obtained by self-mentoring are the same as in Table \ref{tab:acc_caps} \label{tab:acc_caps_large}}}
\end{table}

\subsection{Robustness to covariate shift} 
\label{sub:robustness_to_covariate_shift}
Covariate shift is one of the most difficult situation to handle in machine learning. It corresponds to a gap between the input distributions $p \left( \mathbf{x}  \right) $ of train and test samples while the output conditional distribution given inputs $p \left( \mathbf{y}| \mathbf{x} \right)  $ remains the same. Consequently, a model trained in these conditions will perform much more poorly at test time than what was expected from validation performances obtained at training time. 

There are many possible sources of covariate shift. For example, it is possible that the training set is a rare event w.r.t. the actual data distribution while the likelihood of the data is much higher under a different distribution. Another possibility is that the training set is not independent and identically distributed (iid) w.r.t. the data distribution. It is also possible that the input distribution has simply drifted between the time where training data was acquired and the time where the model is put in production. 

In this subsection, we artificially create one such challenging learning setting by splitting train, validation and test sets in a purposely non-iid way. In the Capsule dataset, due to the Y-shape of the micro-canal, capsule segmentation masks are either located in the middle right side, upper-left side or lower-left side. The split is thus performed based on this positional information as follows:
\begin{itemize}
  \item there are 55 lower-left-side capsules images held-out as test set for evaluation only,
  \item $\mathcal{S}_{\textrm{tr}}$ and $\mathcal{S}_{\textrm{val}}$ contain respectively $n_{\text{s-tr}} \in \left\{ 2;4;5;6;7 \right\}$ and $n_{\text{s-val}}=1$ labeled images of micro-capsules in the middle-right side,
  \item $\mathcal{U}_{\textrm{tr}}$ and $\mathcal{U}_{\textrm{val}}$ contain respectively $n_{\text{s-val}} = 198$ and $n_{\text{s-val}}=20$ unlabeled images of micro-capsules either in the middle-right or in the upper-left side.
\end{itemize}
Observe that in total, we use less images than what the Capsule dataset contains. Indeed, it was necessary to remove some of them in order to create three clearly distinct subsets so as to induce a significant covariate shift. The same patience parameters as in the previous subsection are used. 

Table \ref{tab:acc_caps2} gives average accuracies achieved by the different methods when covariate shift is added. Similarly as in the previous experiment, performances tend to increase with the amount of annotated data. It is also pretty clear that all approaches learn far poorer models with this dataset due to presence of covariate shift. In spite of this, the positive effect of self-mentoring is confirmed by this second setting. Compared to standalone U-net, self-mentoring achieves accuracy increments ranging from $+17.59\%$ up to $+28.83\%$ which are again very significant performance gaps. 
It can remarked that self-training proves to be beneficial in this second experiment with covariate shift but to a lesser extent than self-mentoring. 
Even if self-training is used here for comparison with recent literature, in practice, it is possible to use them jointly since they rely on different but compatible mechanisms. 
Since this task is more challenging than the first one, the protocol starts struggling to converge when $n_{\textrm{s-tr}}+n_{\textrm{s-val}} \leq 4$.

\begin{table}[!h]
\begin{center}
\resizebox{\columnwidth}{!}{
 \begin{tabular}{ c  c | c  c  c }
 \hline
    $n_{\textrm{s-tr}}+n_{\textrm{s-val}}$ & $\lambda_{\textrm{ae}}$ & U-net & Self-trained &  Self-mentored\\
     & & & U-net &  U-net (ours) \\
 \hline
   3 & 10 & 14.31\% & 16.11\% & 36.75\% \\
   5 & 5 & 29.36\% & 30.35\% & 46.95\% \\
   6 & 10 & 28.33\% & 37.28\% & 55.61\% \\
   7 & 5 & 34.49\% & 34.18\% & 58.69\% \\
   8 & 5 & 23.53\% & 30.88\% & 52.36\% \\
  \hline 
 \end{tabular}}
 \end{center}
 \vspace{-0.5em}
 \caption{Segmentation accuracy (Jaccard Indices) with and without self-supervision on the Capsule dataset with covariate shift. \label{tab:acc_caps2}}
\end{table}


\subsection{Coupling with data augmentation} 
\label{sub:coupling_with_data_augmentation}
As previously mentioned, our self-mentoring pipeline works for a predefined number of labeled images but this does not mean that it cannot be combined with DA so that some of those labeled images are actually obtained from DA. 
We show in this subsection that this allows to train U-net for the Capsule dataset under covariate shift with less supervision than what appeared to be necessary in the previous section. 
Before performing DA, we thus start with the same conditions, meaning that we have access to $n_{\textrm{u-tr}}=198 $ and $n_{\textrm{u-val}}=20 $ unlabeled images, $n_{\textrm{s-val}}=1 $ labeled image for validation. Regarding the number of labeled images for training, we will focus on the $n_{\textrm{s-tr}}=2 $ case \modif{because, among previously examined experiments, it is the most difficult situation to cope with. Since DA is meant to boost self-mentoring, it is preferable to test it in this harder setting.}

In this experiment, we use a DA generator to obtain "augmented" images for both the supervised and unsupervised data. The DA generator works as follows:
\begin{enumerate}
  \item sample uniformly at random a rotation + flipping. The possible rotation angles are $0, 90, 180$ and $270$ degrees and the flipping is either horizontal or omitted. Each combination (out of eight possible) has equal probability to occur, including $0$ degree rotation and absence of flipping which leaves the images unchanged. In addition, the chosen modification is performed only with probability $0.5$, otherwise the image is left unchanged.
  \item sample uniformly at random a noise type from the following four categories: no noise, salt and pepper, centered Gaussian (with 0.15 std deviation) and uniform in $\left[ -\frac{1}{2}; \frac{1}{2} \right] $. For the salt and pepper noise, $5\%$ of the pixels are modified. 
\end{enumerate}

Based on the above description, the DA generator does not modify the input image $\mathbf{x}$ with probability $\frac{9}{64}\approx 0.14$. This generator is used in steps 2. to 4. of the pipeline. This means that the losses \eqref{eq:loss_sup}, \eqref{eq:loss_rev}, \eqref{eq:loss_cons} and \eqref{eq:loss_ae} are now computed from "augmented" datasets denoted respectively by $\textrm{DA} \left( \mathcal{S}_{\textrm{tr}} \right) $ and $\textrm{DA} \left( \mathcal{U}_{\textrm{tr}} \right) $. 
Note that we do not use DA on validation sets $\mathcal{S}_{\textrm{val}} $ and $\mathcal{U}_{\textrm{val}} $. 
The generator is used after each epoch to renew the datasets. The size of $\textrm{DA} \left( \mathcal{S}_{\textrm{tr}} \right) $ and $\textrm{DA} \left( \mathcal{U}_{\textrm{tr}} \right) $ is set to 100 images. 
When creating an augmented augmented pair $\left( \mathbf{x}, \mathbf{y} \right) \in \textrm{DA} \left( \mathcal{S}_{\textrm{tr}} \right)$, the augmented input $\mathbf{x}$ is obtained by applying both steps a) and b) but the augmented output $\mathbf{y}$ is obtained by applying a) only. 
For augmented inputs in $\textrm{DA} \left( \mathcal{U}_{\textrm{tr}} \right)$, we use step a) only because adding noise will make the optimization of $L_{\textrm{ae}}$ too difficult. 

\begin{table}[!h]
\begin{center}
\resizebox{\columnwidth}{!}{
 \begin{tabular}{ c  c  c  c }
 \hline
     U-net & U-net  &  Self-mentored  &  Self-mentored  \\
      & + DA &   U-net + DA on $\mathcal{S}_{\textrm{tr}}$ &   U-net + full DA \\     
 \hline
   14.31\% & 40.27\% & 67.39\% & 65.55\% \\
   & & ($\lambda_{\textrm{ae}}=20$) & ($\lambda_{\textrm{ae}}=5$) \\
  \hline 
 \end{tabular}}
 \end{center}
 \vspace{-0.5em}
 \caption{Segmentation accuracy (Jaccard Indices) with and without DA / self-supervision on the Capsule dataset with $n_{\textrm{s-tr}}+n_{\textrm{s-val}}=3$ labeled images. \label{tab:acc_caps_DA}}
\end{table}

JI accuracies are reported in Table \ref{tab:acc_caps_DA} when the protocol is launched with patience set to 50 for steps 2. and 3. and to 100 for the final step. We investigate two DA strategies: either using DA on $\mathcal{S}_{\textrm{tr}}$ only or using DA on $\mathcal{S}_{\textrm{tr}}$ and $\mathcal{U}_{\textrm{tr}}$ (full DA). 
Two important conclusions arise from these results. The first one is that DA alone does not deliver an accuracy increment as high as our self-mentoring pipeline. The second (and more important) conclusion is that DA and our approach contribute to improve accuracy in different ways which is why using them both gives even better results. 
The combined approach achieves far better result on the three-shot learning task than without DA (see first row of Table \ref{tab:acc_caps2}). 
It also seems that it is sufficient to deploy DA on $\mathcal{S}_{\textrm{tr}}$ only as using it also on $\mathcal{U}_{\textrm{tr}}$ does not provide improved performances.
For even more severe FSL settings, the increment between self-mentoring and self-mentoring+DA is smaller. 
It seems that, in this case, DA cannot compensate enough for the lack of supervision in the original data. 


\subsection{Transferability of the referee} 
\label{sub:transferability_of_the_referee}

Although this paper is focused on micro-capsule image analysis, reliably assessing the transferability of the referee network requires using another image dataset. 
The PhC-C2DH-U373 dataset\footnote{\modif{Data provided by Dr. S. Kumar. Department of Bio-
engineering, University of California at Berkeley, Berkeley CA (USA)).}} contains images of Glioblastoma astrocytoma U373 cells filmed on a polyacrylamide substrate. In this film, five cells are present inside each image and the dataset contains binary masks for each of them. 
\modif{This dataset contains 115 images of resolution $512 \times 512$ pixels. }
Experiments on this dataset are meant to illustrate the re-usability of the trained referee on another segmentation task but they also allow to test the protocol in different conditions where, in particular, the objects to segment are not transparent anymore as opposed to the Capsule dataset. 

\begin{figure*}
     \centering
         \includegraphics[width=.19\textwidth]{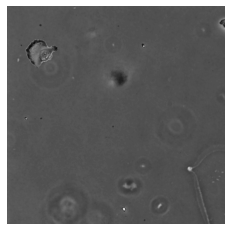}\:
         \includegraphics[width=.19\textwidth]{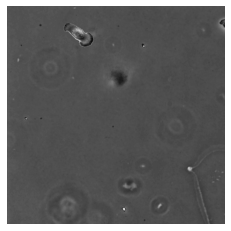}\:
         \includegraphics[width=.19\textwidth]{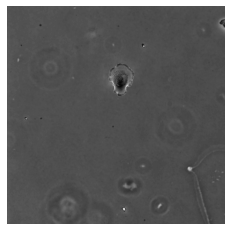}\:
         \includegraphics[width=.19\textwidth]{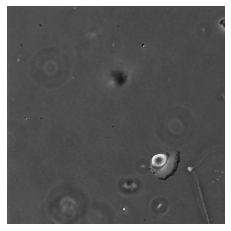}\:
         \includegraphics[width=.19\textwidth]{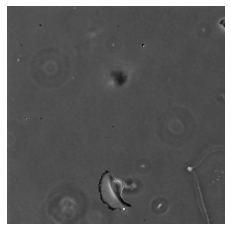}\\
         \includegraphics[width=.19\textwidth]{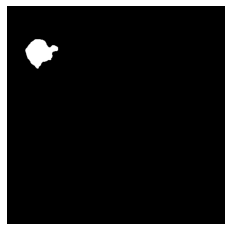}\:
         \includegraphics[width=.19\textwidth]{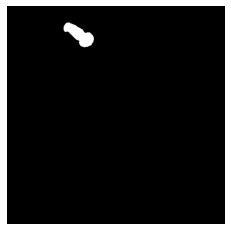}\:
         \includegraphics[width=.19\textwidth]{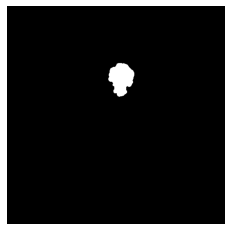}\:
         \includegraphics[width=.19\textwidth]{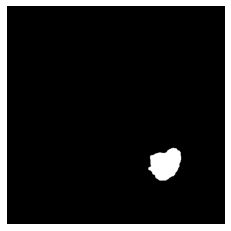}\:
         \includegraphics[width=.19\textwidth]{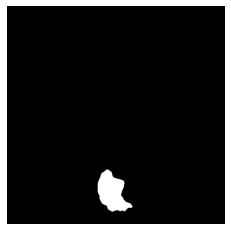}\:

        \caption{Image example for each of the five cells of the PhC-C2DH-U373 dataset. Input images (first row), cell masks $\mathbf{y}$ (second row).}
        \label{fig:cells} 

\end{figure*}

To evaluate the transferability of the referee, it is however necessary to place ourselves in the setting for which it was trained: correcting the mask of a single object. Consequently, a synthetic version of the dataset was created to comply to this setting. We first compute the median image $\mathbf{b}$ of the 115 images contained in the original dataset which yields a fairly good approximation of the background of those images. Then for each of the five cells and each original film image $\mathbf{o}$, we create one artificial input image $\mathbf{x}$ from a cell mask $\mathbf{y}$ as follows: $\mathbf{x} = \mathbf{y}\odot \mathbf{o} + \left( 1 -  \mathbf{y} \right)\odot  \mathbf{b}$. This procedure is applied for each of the five cells visible in $\mathbf{o}$.
The synthetic version of PhC-C2DH-U373 dataset thus contains $5 \times 115 = 575$ $\left( \mathbf{x}, \mathbf{y} \right) $ pairs. One such training pair for each of the five cells is shown in Fig. \ref{fig:cells}.

We voluntarily place ourselves in a challenging learning situation as $\mathcal{S}_{\text{tr}}$ contains only $n_{\textrm{s-tr}}=10$ labeled images and $n_{\textrm{s-val}}=5$ other annotated images are used for ES. These labeled images are spanned from the same cell. $\mathcal{U}_{\text{tr}}$ contains 325 unlabeled images and $\mathcal{U}_{\text{val}}$ contains 20 images. These unlabeled images are spanned from three cells in total, one of which is the same cell as the one used for annotated images. 
Finally, test samples are obtained from the remaining 230 images spanned by the two remaining cells. Similarly as in the previous subsection, splitting the dataset in this way creates a form of covariate shift as test samples contain images of cells that were never seen at training time. In particular, those two "test cells" are located in image regions where no cell appears in training or validation images. This means that if a network overfits by learning cell positions instead of cell patterns it will perform terribly on the test set. 

In spite of this far-from-ideal learning framework, the proposed learning scenario is able to improve the segmentation performance by a large margin. With a slightly bigger U-net model than with the Capsule dataset ($F=10$), standalone U-net achieves $26.45\%$ JI-accuracy while self-mentored U-net achieves $81.16\%$ JI-accuracy. \modif{A large U-net enjoying as many parameters as the cumulated number of parameters used by all networks as part of self-mentoring achieves $73.16\%$ JI-accuracy. }
The reported JI-accuracies are averaged from five different runs where the assignment of a cell to the train, validation or test set is changed. 
This significant gain in accuracy was obtained by re-using the referee that was trained as part of the experiments on the Capsule dataset presented in the previous sub-sections.
\modif{In the original version of the PhC-C2DH-U373 dataset, a large U-net achieves $92.03\%$ JI-accuracy when trained in the fully supervised setting \cite{ronneberger2015u}. Although the version of the dataset we use in this paper is more difficult because of the presence of covariate shift and direct comparisons are not possible, this performance can be seen as an upper bound. It is remarkable that self-mentoring achieves 88\% of this bound from only 15 annotated images.}

\section{Discussion} 
\label{sec:discussion}



This paper introduces a self-supervision scenario, called self-mentoring, to train efficiently a U-net from a very few labeled data and moderately many unlabeled data. This training protocol uses U-net instances called trainee, referee and reverse networks. The pipeline goes as follows: the trainee solves the main segmentation task and maps input images to segmentation masks, the referee delivers a correction to the predicted masks issued by the trainee and finally the reverse network maps output masks produced by the referee back to input images. 
The referee network is trained on purely synthetic data which allows to distill its knowledge to guide the trainee into producing better predictions. 
The reverse net allows to use unlabeled data as part as of an auto-encoding reconstruction loss. 

This contribution is validated on two different segmentation tasks where remarkable accuracy increments are achieved thanks to the self-supervision spanned by the proposed training scenario. The present contribution is able to procure such self-supervision when input images contain only one object to segment on a relatively uniform background. 
The performance improvements are comparable to those obtained from data augmentation alone\footnote{This remark holds for the tested DA in our experiments but other DA implementations may achieve higher performances.} and when the approach is combined with data augmentation, significantly higher improvements are achieved so that a model can be learnt from an extremely small number of annotated images. Also, the approach is compared to its closest alternative in the recent literature known as self-training \cite{xie2020self}.

A computation overhead is obviously incurred by using the proposed self-mentoring pipeline. Compared to a situation where a single U-net has to be trained once, self-mentoring requires the training of the referee, the training of the reverse net and several trainings of the trainee. 
As an indication, we provide the following computation load details for the experiment on the Capsule dataset from \ref{sub:validation_of_the_pipeline_on_the_capsule_dataset}:


\begin{itemize}
  \item the model that is the most time-consuming to learn is the referee network. It uses more filters ($F=30$), more data per epoch (300) and more epochs (from 2500 to 3000).
But as shown from subsection \ref{sub:transferability_of_the_referee}, the referee exhibits a good level and transferability which is perhaps even more important in terms of energy consumption throughout its life-cycle. 

 \item the trainee pre-training is achieved in around 80 epochs in average while the reverse net uses around 40 epochs. 

 \item across the CL loop of the main training phase, several training runs of the returned U-net model are necessary. At the first iteration, a convergence is reached in around 100 epochs. In the last CL iteration, a convergence is reached in around 50 epochs. This number decreases because after each iteration, the model is more and more trained and thus will reach a local minimum of the loss more rapidly. 
\end{itemize}

ES creates a form of randomness in the number of epochs which makes it difficult to obtain an accurate comparison of run times of U-net alone versus self-mentored U-net. 
By fixing the number of epochs to some number $\tau$ for each training in phase 2. to 4. and by setting aside the referee which can be re-used for different datasets, it becomes possible to draw comparisons. If $N$ denotes the number of CL steps, the overhead is approximately\footnote{We neglect the time spent on computing validation losses for ES.} given by $\tau \left[    \left( N+1 \right) \alpha_1n_{\textrm{s-tr}} + N \alpha_2  n_{\textrm{u-tr}}  \right]$. The constant $\alpha_1$ denotes the processing time for a single datum in the reverse net or the trainee (since they share the same architecture) while $\alpha_2 > \alpha_1$ denotes that of the combined networks. The first term comes from the optimization of losses $L_{\text{sup}}$ and $L_{\text{sup}}^{(-1)}$ while the second term comes from the optimization of losses $L_{\text{ae}}$ and $L_{\text{cons}}$. In our semi-supervised FSL context, the dominant term is the second one since we have a lot more unsupervised images than supervised ones.

Besides, the incurred computation cost overhead has to be contrasted with the performance increment obtained by the protocol. Indeed, training a U-net in a more usual supervised setting involves way more annotated data and longer training time for U-net alone which would make the ratio far smaller. But in turn, as the number for annotated data gets higher, the added value of the protocol is reduced. 
Consequently, self-mentoring should be used when lighter options are doomed to fail, i.e. typically in very few shot learning contexts. 
Finally, it should be noted that the returned U-net is no bigger than the U-net trained alone on the amount of available annotated data. This means that, at inference time, no overhead is incurred. 

From a practical standpoint, the main limitation of the proposed self-mentoring pipeline is that it is designed for semantic segmentation of a single object. Future works will focus on an extension to instance segmentation where multiple objects can be present in images. For that purpose, it is planned to use the protocol with a different base model from U-net. We believe that mask R-CNN \cite{he2017mask} could be an alternative allowing the protocol to work also for instance segmentation. Mask R-CNN achieves impressive instance segmentation results in complex scenes. Swapping mask R-CNN and U-net is, however, not immediate. Note that, by mapping the output of R-CNN to multiple single-object masks, this batch of masks could still be analyzed by the same referee as the one introduced in this paper.

Finally, another strong hypothesis on which the protocol relies is the homogeneity of the image background. While this is less crucial in microscopy datasets where backgrounds are very homogeneous, it is still desirable to secure more robust versions of the protocol in case of increased background variability. This will require to use an additional network to encode the background style which is a necessary input for the reverse network to produce the appropriate background from a binary mask.

\section{Conclusion}
A new deep learning pipeline, called self-mentoring, has been introduced to segment bio-artificial capsules in microscopy images. The pipeline mainly leverages the simple structure of segmentation mask images for such objects. By training a referee network to produce correct mask regions from imperfect ones, this paper proves that the referee network can mentor a trainee network to solve the main segmentation task. An important aspect of this pipeline is that the referee can be efficiently trained on a massive synthetic dataset. Thanks to the knowledge transferred from the referee, the trainee can achieve good segmentation performances even if it has access to only a very few annotated training examples. By coupling the proposed pipeline with data augmentation, it becomes possible for bio-mechanical analysts to obtain bio-artificial capsule segments of an entire microscopy film by only providing a handful of annotated images.


\section*{Acknowledgment}

This work was performed using HPC resources from GENCI-IDRIS (Grant 2021-AD011011606R1).


\bibliographystyle{plain}
\bibliography{ref}

\end{document}